\documentclass[runningheads]{llncs}

\usepackage[mobile]{eccv}

\usepackage{eccvabbrv}

\usepackage{graphicx}
\usepackage{booktabs}
\usepackage{amsmath}
\usepackage{amssymb}
\usepackage{color}
\usepackage{colortbl}
\usepackage{makecell}
\usepackage{multicol}
\usepackage{multirow}
\usepackage{pifont}
\usepackage{marvosym}

\newcommand{\cmark}{\ding{51}}
\newcommand{\xmark}{\ding{55}}

\usepackage{tabularx}
\usepackage{tikz}
\usepackage{wrapfig}

\definecolor{sf_blue}{RGB}{66, 133, 244}
\definecolor{sf_red}{RGB}{231, 66, 52}
\definecolor{sf_yellow}{RGB}{251, 189, 5}
\definecolor{sf_green}{RGB}{51, 168, 82}
\definecolor{sf_gray}{RGB}{165, 165, 165}

\definecolor{nu_barrier}{RGB}{112, 128, 144}
\definecolor{nu_bicycle}{RGB}{220, 20, 60}
\definecolor{nu_bus}{RGB}{255, 0, 0}
\definecolor{nu_car}{RGB}{255, 158, 0}
\definecolor{nu_cons}{RGB}{233, 150, 70}
\definecolor{nu_motor}{RGB}{255, 61, 99}
\definecolor{nu_ped}{RGB}{0, 0, 230}
\definecolor{nu_cone}{RGB}{47, 79, 79}
\definecolor{nu_trailer}{RGB}{255, 140, 0}
\definecolor{nu_truck}{RGB}{255, 99, 71}
\definecolor{nu_driv}{RGB}{0, 207, 191}
\definecolor{nu_flat}{RGB}{175, 0, 75}
\definecolor{nu_sidewalk}{RGB}{75, 0, 75}
\definecolor{nu_terrain}{RGB}{112, 180, 60}
\definecolor{nu_manmade}{RGB}{222, 184, 135}
\definecolor{nu_veg}{RGB}{0, 175, 0}

\usepackage[accsupp]{axessibility}
\usepackage[pagebackref,breaklinks,colorlinks,citecolor=eccvblue,linkcolor=eccvblue]{hyperref}

\usepackage{orcidlink}

\begin{document}

\title{4D Contrastive Superflows are Dense 3D Representation Learners}

\titlerunning{4D Contrastive Superflows are Dense 3D Representation Learners}

\author{
Xiang Xu\inst{1,}\thanks{X. Xu and L. Kong contributed equally to this work. \textrm{\Letter} Corresponding author.}
\and
Lingdong Kong\inst{2,3,*}
\and
Hui Shuai\inst{4}
\and
Wenwei Zhang\inst{2}
\and\\
Liang Pan\inst{2}
\and
Kai Chen\inst{2}
\and
Ziwei Liu\inst{5}
\and
Qingshan Liu\inst{4,\textrm{\Letter}}
}

\authorrunning{X.~Xu et al.}

\institute{
Nanjing University of Aeronautics and Astronautics
\and
Shanghai AI Laboratory
\and
National University of Singapore
\and
Nanjing University of Posts and Telecommunications
\and
S-Lab, Nanyang Technological University
}

\maketitle

\begin{abstract}
  In the realm of autonomous driving, accurate 3D perception is the foundation. However, developing such models relies on extensive human annotations -- a process that is both costly and labor-intensive. To address this challenge from a data representation learning perspective, we introduce \textbf{SuperFlow}, a novel framework designed to harness consecutive LiDAR-camera pairs for establishing spatiotemporal pretraining objectives. SuperFlow stands out by integrating two key designs: \textbf{1)} a dense-to-sparse consistency regularization, which promotes insensitivity to point cloud density variations during feature learning, and \textbf{2)} a flow-based contrastive learning module, carefully crafted to extract meaningful temporal cues from readily available sensor calibrations. To further boost learning efficiency, we incorporate a plug-and-play view consistency module that enhances the alignment of the knowledge distilled from camera views. Extensive comparative and ablation studies across 11 heterogeneous LiDAR datasets validate our effectiveness and superiority. Additionally, we observe several interesting emerging properties by scaling up the 2D and 3D backbones during pretraining, shedding light on the future research of 3D foundation models for LiDAR-based perception. Code is publicly available at \url{https://github.com/Xiangxu-0103/SuperFlow}.

\end{abstract}

\section{Introduction}
\label{sec:introduction}

Driving perception is one of the most crucial components of an autonomous vehicle system. Recent advancements in sensing technologies, such as light detection and ranging (LiDAR) sensors and surrounding-view cameras, open up new possibilities for a holistic, accurate, and 3D-aware scene perception \cite{badue2021survey,rizzoli2022survey,cao2024pasco}.

Training a 3D perception model that can perform well in real-world scenarios often requires large-scale datasets and sufficient computing power \cite{gao2021survey,liu2024survey}. Different from 2D, annotating 3D data is notably more expensive and labor-intensive, which hinders the scalability of existing 3D perception models \cite{muhammad2020survey,xiao2023survey,zhang2023survey,geiger2012kitti}. Data representation learning serves as a potential solution to mitigate such a problem \cite{bengio2013survey,puy2024three}. By designing suitable pretraining objectives, the models are anticipated to extract useful concepts from raw data, where such concepts can help improve models' performance on downstream tasks with fewer annotations \cite{lekhac2020survey}.

\begin{wrapfigure}{r}{0.56\textwidth}
    \vspace{-0.65cm}
    \centering
    \includegraphics[width=0.56\textwidth]{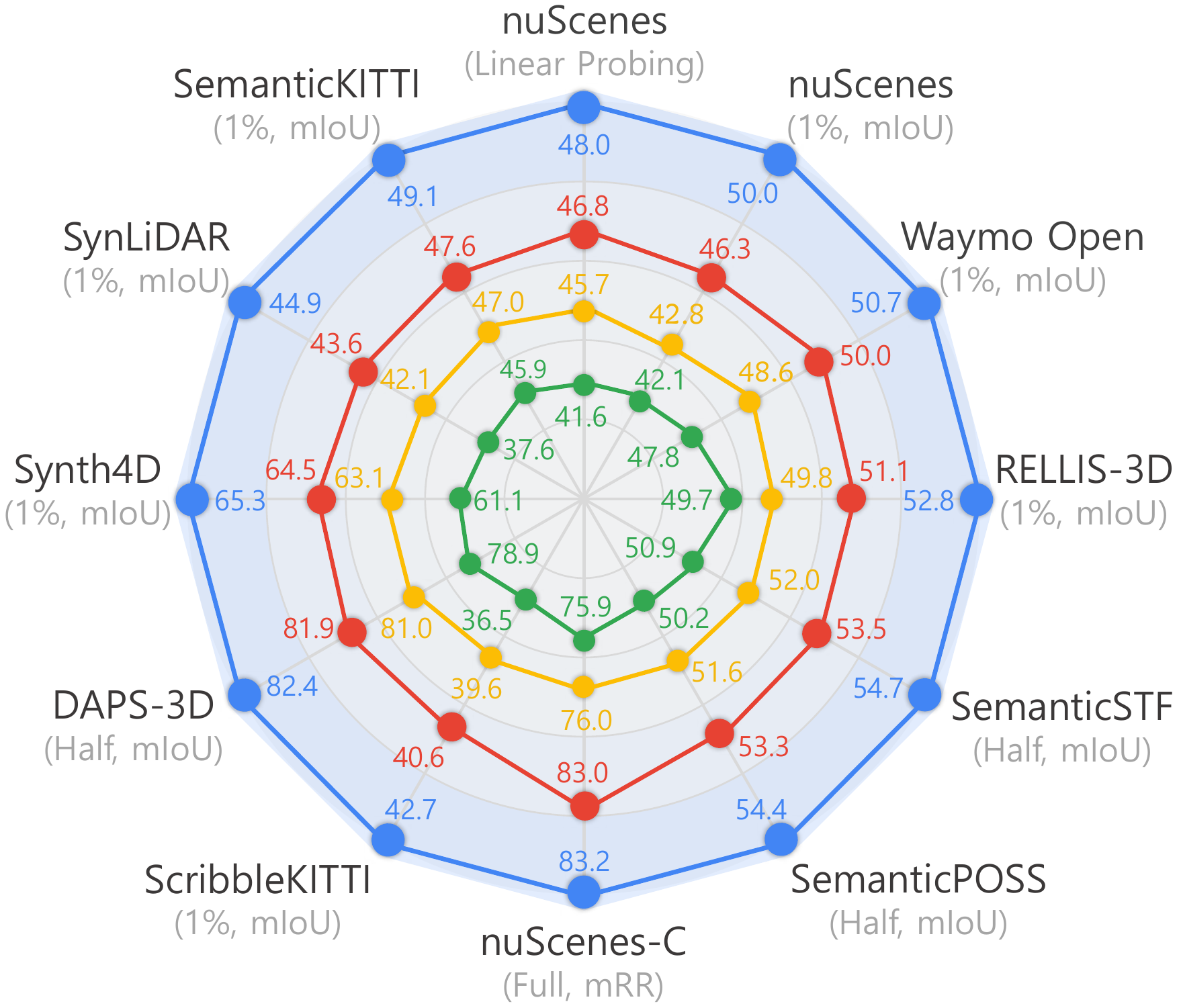}
    \vspace{-0.55cm}
    \caption{\textbf{Performance overview} of \textcolor{sf_blue}{SuperFlow} compared to state-of-the-art image-to-LiDAR pretraining methods, \ie, \textcolor{sf_red}{Seal} \cite{seal}, \textcolor{sf_yellow}{SLidR} \cite{slidr}, and \textcolor{sf_green}{PPKT} \cite{ppkt}, on eleven LiDAR datasets. The scores of prior methods are normalized based on SuperFlow's scores. The larger the area coverage, the better the overall segmentation performance.}
    \label{fig:teaser}
    \vspace{-1cm}
\end{wrapfigure}

Recently, Sautier \etal \cite{slidr} proposed SLidR to distill knowledge from surrounding camera views -- using a pretrained 2D backbone such as MoCo \cite{chen2020moCoV2} and DINO \cite{oquab2023dinov2} -- to LiDAR point clouds, exhibiting promising 3D representation learning properties. The key to its success is the superpixel-driven contrastive objectives between cameras and LiDAR sensors. Subsequent works further extended this framework from various aspects, such as class balancing \cite{st-slidr}, hybrid-view distillation \cite{zhang2024hvdistill}, semantic superpixels \cite{seal,2023CLIP2Scene,chen2023towards}, and so on. While these methods showed improved performance over their baselines, there exist several issues that could undermine the data representation learning.

The first concern revolves around the inherent temporal dynamics of LiDAR data \cite{nuScenes,behley2021semanticKITTI}. LiDAR point clouds are acquired sequentially, capturing the essence of motion within the scene. Traditional approaches \cite{ppkt,slidr,st-slidr,zhang2024hvdistill,seal} often overlook this temporal aspect, treating each snapshot as an isolated scan. However, this sequential nature holds a wealth of information that can significantly enrich the model's understanding of the 3D environment \cite{nunes2023tarl,wu2023stssl}. Utilizing these temporal cues can lead to more robust and context-aware 3D perception models, which is crucial for dynamic environments encountered in autonomous driving.

Moreover, the varying density of LiDAR point clouds presents a unique challenge \cite{kong2023robo3D,kong2022lasermix,unal2022scribbleKITTI}. Due to the nature of LiDAR scanning and data acquisition, different areas within the same scene can have significantly different point densities, which can in turn affect the consistency of feature representation across the scene \cite{aygun2021pls4d,zhou2020polarNet,kong2022lasermix,yin2022proposalcontrast}. Therefore, a model that can learn invariant features regardless of point cloud density tends to be effective for recognizing the structural and semantic information in the 3D space.

In lieu of existing challenges, we propose a novel spatiotemporal contrastive learning dubbed \textbf{SuperFlow} to encourage effective cross-sensor knowledge distillation. Our approach features three key components, all centered around the use of the off-the-shelf temporal cues inherent in the LiDAR acquisition process:
\begin{itemize}
    \item We first introduce a straightforward yet effective view consistency alignment that seamlessly generates semantic superpixels with language guidance, alleviating the ``self-conflict'' issues in existing works \cite{slidr,st-slidr,seal}. As opposed to the previous pipeline, our method also aligns the semantics across camera views in consecutive scenes, paving the way for more sophisticated designs.
    \item To address the varying density of LiDAR point clouds, we present a dense-to-sparse regularization module that encourages consistency between features of dense and sparse point clouds. Dense points are obtained by concatenating multi-sweep LiDAR scans within a suitable time window and propagating the semantic superpixels from sparse to dense points. By leveraging dense point features to regularize sparse point features, the model promotes insensitivity to point cloud density variations.
    \item To capture useful temporal cues from consecutive scans across different timestamps, we design a flow-based contrastive learning module. This module takes multiple LiDAR-camera pairs as input and excites strong consistency between temporally shifted representations. Analogous to existing image-to-LiDAR representation learning methods \cite{slidr,st-slidr,seal}, we also incorporate useful spatial contrastive objectives into our framework, setting a unified pipeline that emphasizes holistic representation learning from both the structural 3D layouts and the temporal 4D information.
\end{itemize}

The strong spatiotemporal consistency regularization in SuperFlow effectively forms a semantically rich landscape that enhances data representations. As illustrated in \cref{fig:teaser}, our approach achieves appealing performance gains over state-of-the-art 3D pretraining methods across a diverse spectrum of downstream tasks. Meanwhile, we also target at scaling the capacity of both 2D and 3D backbones during pretraining, shedding light on the future development of more robust, unified, and ubiquitous 3D perception models.

To summarize, this work incorporates key contributions listed as follows:
\begin{itemize}
    \item We present \textbf{SuperFlow}, a novel framework aimed to harness consecutive LiDAR-camera pairs for establishing spatiotemporal pretraining objectives.
    \item Our framework incorporates novel designs including view consistency alignment, dense-to-sparse regularization, and flow-based contrastive learning, which better encourages data representation learning effects between camera and LiDAR sensors across consecutive scans.
    \item Our approach sets a new state-of-the-art performance across 11 LiDAR datasets, exhibiting strong robustness and generalizability. We also reveal intriguing emergent properties as we scale up the 2D and 3D backbones, which could lay the foundation for scalable 3D perception.
\end{itemize}

\section{Related Work}
\label{sec:related_work}

\noindent\textbf{LiDAR-based 3D Perception.}
The LiDAR sensor has been widely used in today's 3D perception systems, credited to its robust and structural sensing abilities~\cite{triess2021survey,behley2021semanticKITTI,sun2024mmdet3d-lidarseg}. Due to the sparse and unordered nature of LiDAR point clouds, suitable rasterization strategies are needed to convert them into structural inputs~\cite{uecker2022analyzing,hu2021sensatUrban}. Popular choices include sparse voxels~\cite{choy2019minkowski,zhu2021cylindrical,tang2020searching,hong2021dsnet,hong20224dDSNet,cheng2021af2S3Net}, bird's eye view maps~\cite{zhou2020polarNet,chen2021polarStream,zhou2021panoptic,liong2020amvNet}, range view images~\cite{milioto2019rangenet++,cortinhal2020salsanext,xu2020squeezesegv3,zhao2021fidnet,cheng2022cenet,kong2023rethinking,xu2023frnet}, and multi-view fusion~\cite{xu2021rpvnet,cheng2021af2S3Net,liu2023uniseg,jaritz2020xMUDA,qiu2022gfNet,xu2023multiview,liu2024m3net}. While witnessing record-breaking performances on standard benchmarks, existing approaches rely heavily on human annotations, which hinders scalability~\cite{gao2021survey}. In response to this challenge, we resort to newly appeared 3D representation learning, hoping to leverage the rich collections of unlabeled LiDAR point clouds for more effective learning from LiDAR data. This could further enrich the efficacy of LiDAR-based perception.

\noindent\textbf{Data-Efficient 3D Perception.}
To better save annotation budgets, previous efforts seek 3D perception in a data-efficient manner~\cite{gao2021survey,kong2023conDA,jaritz2020xMUDA,kong2024lasermix2,2023CLIP2Scene,chen2023towards}. One line of research resorts to weak supervision, \textit{e.g.}, seeding points~\cite{shi2022weak,hu2022sqn,li2022coarse3D,zhang2023growsp}, active prompts~\cite{hu2022inter,liu2022less,xie2023annotator}, and scribbles~\cite{unal2022scribbleKITTI}, for weakly-supervised LiDAR semantic segmentation. Another line of research seeks semi-supervised learning approaches~\cite{MeanTeacher,kong2022lasermix,li2023lim3d} to better tackle efficient 3D scene perception and achieve promising results. In this work, different from the prior pursuits, we tackle efficient 3D perception from the data representation learning perspective. We establish several LiDAR-based data representation learning settings that seamlessly combine pretraining with weakly- and semi-supervised learning, further enhancing the scalability of 3D perception systems.

\noindent\textbf{3D Representation Learning.}
Analog to 2D representation learning strategies~\cite{chen2020simCLR,he2020moCo,chen2021moCoV3,xie2022simMIM,he2022mae}, prior works designed contrastive~\cite{pointcontrast,depthcontrast,yin2022proposalcontrast,sautier2023bevcontrast,hou2021exploring,nunes2022segcontrast}, masked modeling~\cite{hess2022masked,krispel2024maeli,wei2023tmae}, and reconstruction~\cite{michele2023saluda,boulch2023also} objectives for 3D pretraining. Most early 3D representation learning approaches use a single modality for pretraining, leaving room for further development. The off-the-shelf calibrations among different types of sensors provide a promising solution for building pretraining objectives~\cite{ppkt}. Recently, SLidR~\cite{slidr} has made the first contribution toward multi-modal 3D representation learning between camera and LiDAR sensors. Subsequent works~\cite{st-slidr,pang2023tricc,zhang2024hvdistill} extended this framework with more advanced designs. Seal~\cite{seal} leverages powerful vision foundation models~\cite{kirillov2023sam,zou2023seem,zhang2023openSeeD,zou2023xdecoder} to better assist the contrastive learning across sensors. Puy \textit{et al.}~\cite{puy2023revisiting,puy2024three} conducted a comprehensive study on the distillation recipe for better pretraining effects. While these approaches have exhibited better performance than their baselines, they overlooked the rich temporal cues across consecutive scans, which might lead to sub-opt pretraining performance. In this work, we construct dense 3D representation learning objectives using calibrated LiDAR sequences. Our approach encourages the consistency between features from sparse to dense inputs and features across timestamps, yielding superiority over existing endeavors.

\noindent\textbf{4D Representation Learning.}
Leveraging consecutive scans is promising in extracting temporal relations~\cite{aygun2021pls4d,hong20224dDSNet,shi2020spsequencenet,duerr2020lidar}. For point cloud data pretraining, prior works~\cite{sheng2023point,liu2023leaf,shen2023masked,zhang2023complete,chen20224dcontrast} mainly focused on applying 4D cues on object- and human-centric point clouds, which are often small in scale. For large-scale automotive point clouds, STRL~\cite{huang2021strl} learns spatiotemporal data invariance with different spatial augmentations in the point cloud sequence. TARL~\cite{nunes2023tarl} and STSSL~\cite{wu2023stssl} encourage similarities of point clusters in two consecutive frames, where such clusters are obtained by ground removal and clustering algorithms, \textit{i.e.}, RANSAC~\cite{foschler1981ransac}, Patchwork~\cite{lim2021patchwork}, and HDBSCAN~\cite{ester1996dbscan}. BEVContrast~\cite{sautier2023bevcontrast} shares a similar motivation but utilizes BEV maps for contrastive learning, which yields a more effective implementation. The ``one-fits-all'' clustering parameters, however, are often difficult to obtain, hindering existing works. Different from existing methods that use a single modality for 4D representation learning, we propose to leverage LiDAR-camera correspondences and semantic-rich superpixels to establish meaningful multi-modality 4D pretraining objectives.

\section{SuperFlow}
\label{sec:methodology}

In this section, we first revisit the common setups of the camera-to-LiDAR distillation baseline (\cf \cref{sec:preliminary}). We then elaborate on the technical details of SuperFlow, encompassing a straightforward yet effective view consistency alignment (\cf \cref{sec:vca}), a dense-to-sparse consistency regularization (\cf \cref{sec:dense_to_sparse}), and a flow-based spatiotemporal contrastive learning (\cf \cref{sec:spatial_temporal}). The overall pipeline of the proposed SuperFlow framework is depicted in \cref{fig:framework}.

\subsection{Preliminaries}
\label{sec:preliminary}

\noindent\textbf{Problem Definition.} Given a point cloud $\mathcal{P}^{t} = \{\mathbf{p}_{i}^{t}, \mathbf{f}_{i}^{t} | i=1,...,N\}$ with $N$ points captured by a LiDAR sensor at time $t$, where $\mathbf{p}_{i} \in \mathbb{R}^{3}$ denotes the coordinate of the point and $\mathbf{f}_{i} \in \mathbb{R}^{C}$ is the corresponding feature, we aim to transfer knowledge from $M$ surrounding camera images $\mathcal{I}^{t} = \{\mathbf{I}_{i}^{t} | i=1,...,M\}$ into the point cloud. Here, $\mathbf{I}_{i} \in \mathbb{R}^{H \times W \times 3}$ represents an image with height $H$ and width $W$. Prior works \cite{slidr, seal} generate a set of class-agnostic superpixels $\mathcal{X}_{i} = \{\mathbf{X}_{i}^{j} | j = 1,...,V\}$ for each image via the unsupervised SLIC algorithm \cite{achanta2012slic} or the more recent vision foundation models (VFMs) \cite{kirillov2023sam,zou2023xdecoder,zou2023seem}, where $V$ denotes the total number of superpixels. Assuming that the point cloud $\mathcal{P}^{t}$ and images $\mathcal{I}^{t}$ are calibrated, the point cloud $\mathbf{p}_{i} = (x_{i}, y_{i}, z_{i})$ can be then projected to the image plane $(u_{i}, v_{i})$ using the following sensor calibration parameters:
\begin{equation}
    [u_{i}, v_{i}, 1]^{\text{T}} = \frac{1}{z_{i}} \times \Gamma_{K} \times \Gamma_{c \leftarrow l} \times [x_{i}, y_{i}, z_{i}]^{\text{T}}~,
    \label{equ:projection}
\end{equation}
where $\Gamma_{K}$ denotes the camera intrinsic matrix and $\Gamma_{c \leftarrow l}$ is the transformation matrix from LiDAR sensors to surrounding-view cameras. We also obtain a set of superpoints $\mathcal{Y} = \{\mathbf{Y}^{j} | j = 1,...,V\}$ through this projection.

\noindent\textbf{Network Representations}. Let $\mathcal{F}_{\theta_{p}}: \mathbb{R}^{N \times (3+C)} \rightarrow \mathbb{R}^{N \times D}$ be a 3D backbone with trainable parameters $\theta_{p}$, which takes LiDAR points as input and outputs $D$-dimensional point features. Let $\mathcal{G}_{\theta_{i}}: \mathbb{R}^{H \times W \times 3} \rightarrow \mathbb{R}^{\frac{H}{S} \times \frac{W}{S} \times E}$ be an image backbone with pretrained parameters $\theta_{i}$ that takes images as input and outputs $E$-dimensional image features with stride $S$. Let $\mathcal{H}_{\omega_{p}}: \mathbb{R}^{N \times D} \rightarrow \mathbb{R}^{N \times L}$ and $\mathcal{H}_{\omega_{i}}: \mathbb{R}^{\frac{H}{S} \times \frac{W}{S} \times E} \rightarrow \mathbb{R}^{H \times W \times L}$ be linear heads with trainable parameters $\omega_{p}$ and $\omega_{i}$, which project backbone features to $L$-dimensional features with $\ell_2$-normalization and upsample image features to $H \times W$ with bilinear interpolation.

\noindent\textbf{Pretraining Objective.}
The overall objective of image-to-LiDAR representation learning \cite{slidr} is to transfer knowledge from the trained image backbone $\mathcal{G}_{\theta_{i}}$ to the 3D backbone $\mathcal{F}_{\theta_{p}}$. The superpixels $\mathcal{X}_{i}$ generated offline, serve as an intermediate to effectively guide the knowledge transfer process.

\begin{figure}[t]
    \centering
    \begin{subfigure}[b]{0.306\textwidth}
        \centering
        \includegraphics[width=\textwidth]{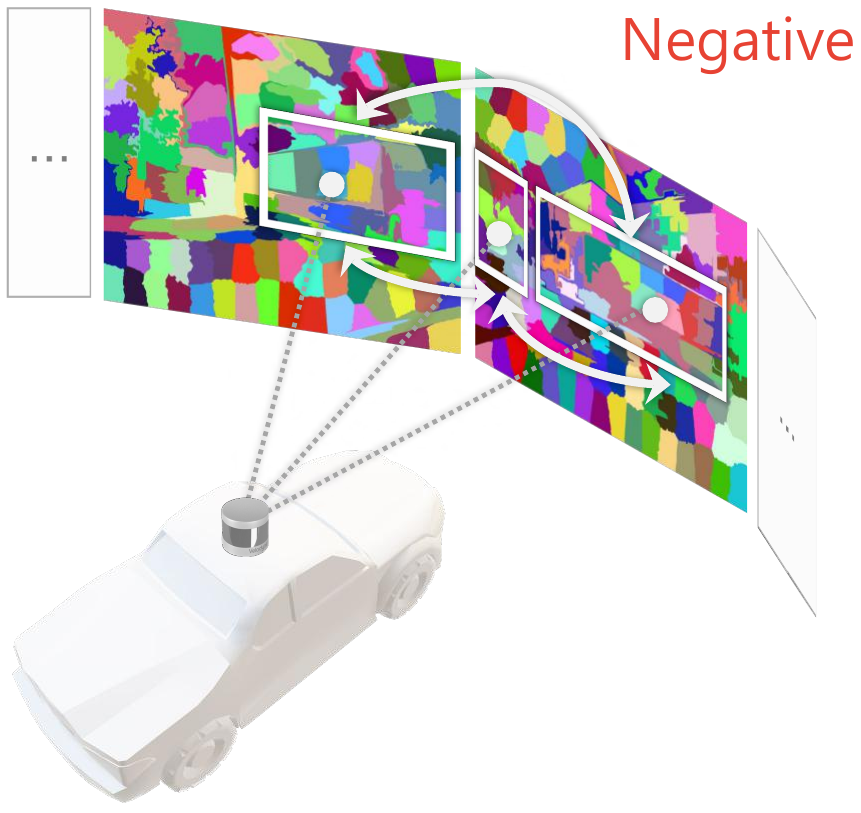}
        \caption{Heuristic}
        \label{fig:slic}
    \end{subfigure}
    ~~
    \begin{subfigure}[b]{0.306\textwidth}
        \centering
        \includegraphics[width=\textwidth]{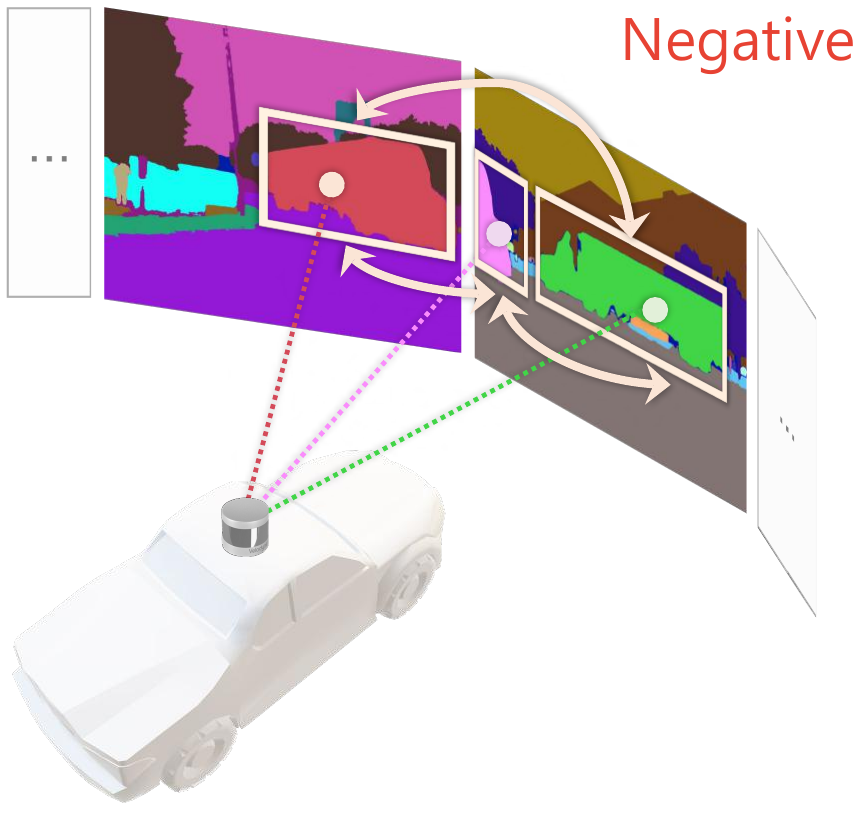}
        \caption{Class Agnostic}
        \label{fig:seem}
    \end{subfigure}
    ~~
    \begin{subfigure}[b]{0.306\textwidth}
        \centering
        \includegraphics[width=\textwidth]{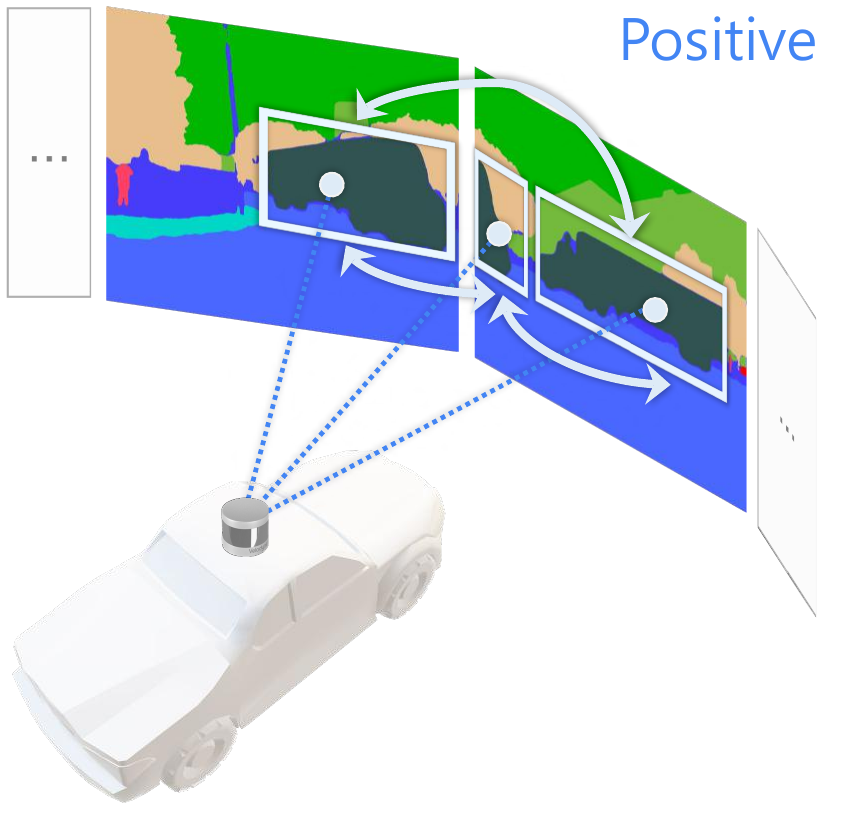}
        \caption{View Consistent}
        \label{fig:vsa}
    \end{subfigure}
    \caption{\textbf{Comparisons of different superpixels.} (a) Class-agnostic superpixels generated by the unsupervised SLIC \cite{achanta2012slic} algorithm. (b) Class-agnostic semantic superpixels generated by vision foundation models (VFMs) \cite{zou2023seem,zhang2023openSeeD,zou2023xdecoder}. (c) View-consistent semantic superpixels generated by our view consistency alignment module.}
    \label{fig:superpixels}
    \vspace{-0.2cm}
\end{figure}

\subsection{View Consistency Alignment}
\label{sec:vca}

\noindent\textbf{Motivation.}
The class-agnostic superpixels $\mathcal{X}_{i}$ used in prior works \cite{slidr,st-slidr,seal} are typically instance-level and do not consider their actual categories. As discussed in \cite{st-slidr}, instance-level superpixels can lead to ``self-conflict'' problems, which undermines the effectiveness of pretraining.

\noindent\textbf{Superpixel Comparisons.}
\cref{fig:superpixels} compares superpixels generated via the unsupervised SLIC \cite{achanta2012slic} and VFMs. SLIC \cite{achanta2012slic} tends to over-segment objects, causing semantic conflicts. VFMs generate superpixels through a panoptic segmentation head, which can still lead to ``self-conflict'' in three conditions (see \cref{fig:seem}): \ding{172} when the same object appears in different camera views, leading to different parts of the same object being treated as negative samples; \ding{173} when objects of the same category within the same camera view are treated as negative samples; \ding{174} when objects across different camera views are treated as negative samples even if they share the same label.

\noindent\textbf{Semantic-Related Superpixels Generation.}
To address these issues, we propose generating semantic-related superpixels to ensure consistency across camera views. Contrastive Vision-Language Pre-training (CLIP) \cite{CLIP} has shown great generalization in few-shot learning. Building on existing VFMs \cite{kirillov2023sam,zou2023xdecoder,zou2023seem}, we employ CLIP's text encoder and fine-tune the last layer of the segmentation head from VFMs with predefined text prompts. This allows the segmentation head to generate language-guided semantic categories for each pixel, which we leverage as superpixels. As shown in \cref{fig:vsa}, we unify superpixels across camera views based on semantic category, alleviating the ``self-conflict'' problem in prior image-to-LiDAR contrastive learning pipelines.

\subsection{D2S: Dense-to-Sparse Consistency Regularization}
\label{sec:dense_to_sparse}

\noindent\textbf{Motivation.}
LiDAR points are sparse and often incomplete, significantly restricting the efficacy of the cross-sensor feature representation learning process. In this work, we propose to tackle this challenge by combining multiple LiDAR scans within a suitable time window to create a dense point cloud, which is then used to encourage consistency with the sparse point cloud.

\begin{wrapfigure}{r}{0.56\textwidth}
    \centering
    \vspace{-0.7cm}
    \includegraphics[width=0.56\textwidth]{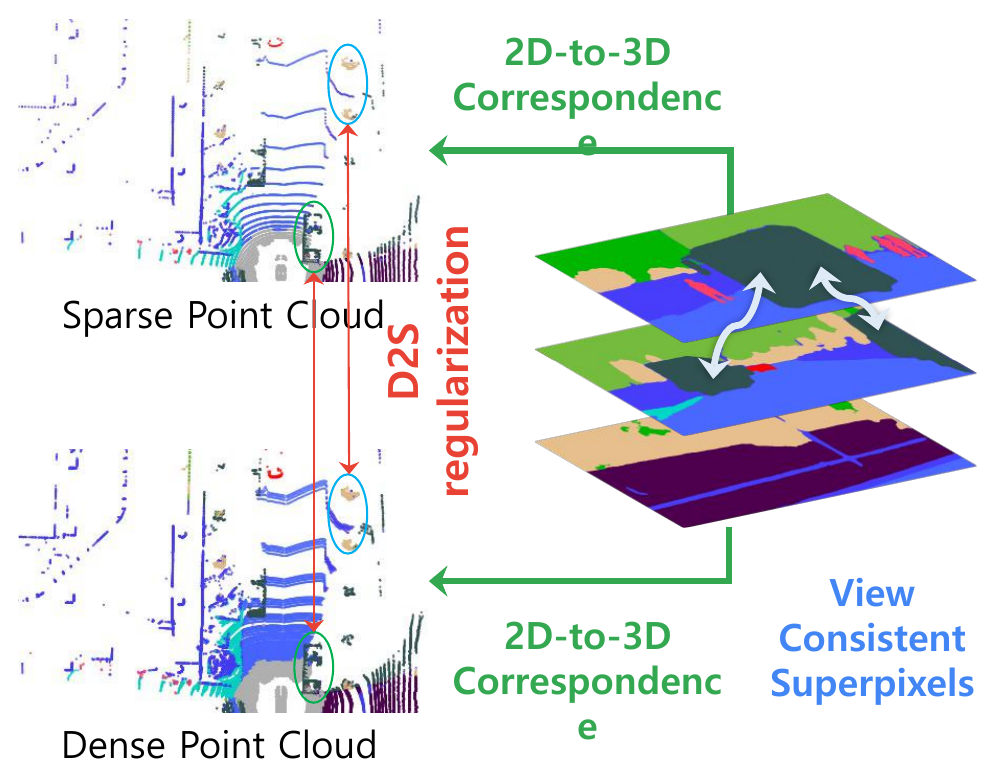}
    \vspace{-0.6cm}
    \caption{\textbf{Dense-to-sparse (D2S)} consistency regularization module. Dense point clouds are obtained by combining multiple point clouds captured at different times. A D2S regularization is formulated by encouraging the consistency between dense features and sparse features.}
    \label{fig:d2s}
    \vspace{-0.5cm}
\end{wrapfigure}

\noindent\textbf{Point Cloud Concatenation.}
Specifically, given a keyframe point cloud $\mathcal{P}^{t}$ captured at time $t$ and a set of sweep point clouds $\{\mathcal{P}^{s} | s = 1,...,T\}$ captured at previous times $s$, we first transform the coordinate $(x^{s}, y^{s}, z^{s})$ of the sweep point cloud $\mathcal{P}^{s}$ to the coordinate systems of $\mathcal{P}^{t}$, as they share different systems due to the vehicle's movement:
\begin{equation}
    [\tilde{x}^{s}, \tilde{y}^{s}, \tilde{z}^{s}]^{\text{T}} = \Gamma_{t \leftarrow s} \times [x^{s}, y^{s}, z^{s}]^{\text{T}}~,
    \label{eq:sweep}
\end{equation}
where $\Gamma_{t \leftarrow s}$ denotes the transformation matrix from the sweep point cloud at time $s$ to the keyframe point cloud at time $t$. We then concatenate the transformed sweep points $\{\tilde{\mathcal{P}}^{s}|s = 1,...T\}$ with $\mathcal{P}^{t}$ to obtain a dense point cloud $\mathcal{P}^{d}$. As shown in \cref{fig:d2s}, $\mathcal{P}^{d}$ fuses temporal information from consecutive point clouds, resulting in a dense and semantically rich representation for feature learning.

\noindent\textbf{Dense Superpoints.}
Meanwhile, we generate sets of superpoints $\mathcal{Y}^{d}$ and $\mathcal{Y}^{t}$ for $\mathcal{P}^{d}$ and $\mathcal{P}^{t}$, respectively, using superpixels $\mathcal{X}^{t}$. Both $\mathcal{P}^{t}$ and $\mathcal{P}^{d}$ are fed into the weight-shared 3D network $\mathcal{F}_{\theta_{p}}$ and $\mathcal{H}_{\omega_{p}}$ for feature extraction. The output features are grouped via average pooling based on the superpoint indices to obtain superpoint features $\mathbf{Q}^{d}$ and $\mathbf{Q}^{t}$, where $\mathbf{Q}^{d} \in \mathbb{R}^{V \times L}$ and $\mathbf{Q}^{d} \in \mathbb{R}^{V \times L}$. We expect $\mathbf{Q}^{d}$ and $\mathbf{Q}^{t}$ to share similar features, leading to the following D2S loss:
\begin{equation}
    \mathcal{L}_{\text{d2s}} = \frac{1}{V}\sum_{i = 1}^{V} (1 - <\mathbf{q}_{i}^{t}, \mathbf{q}_{i}^{d}>)~,
    \label{equ:regularization}
\end{equation}
where $<\cdot, \cdot>$ denotes the scalar product to measure the similarity of features.

\begin{figure}[t]
    \begin{center}
    \includegraphics[width=\textwidth]{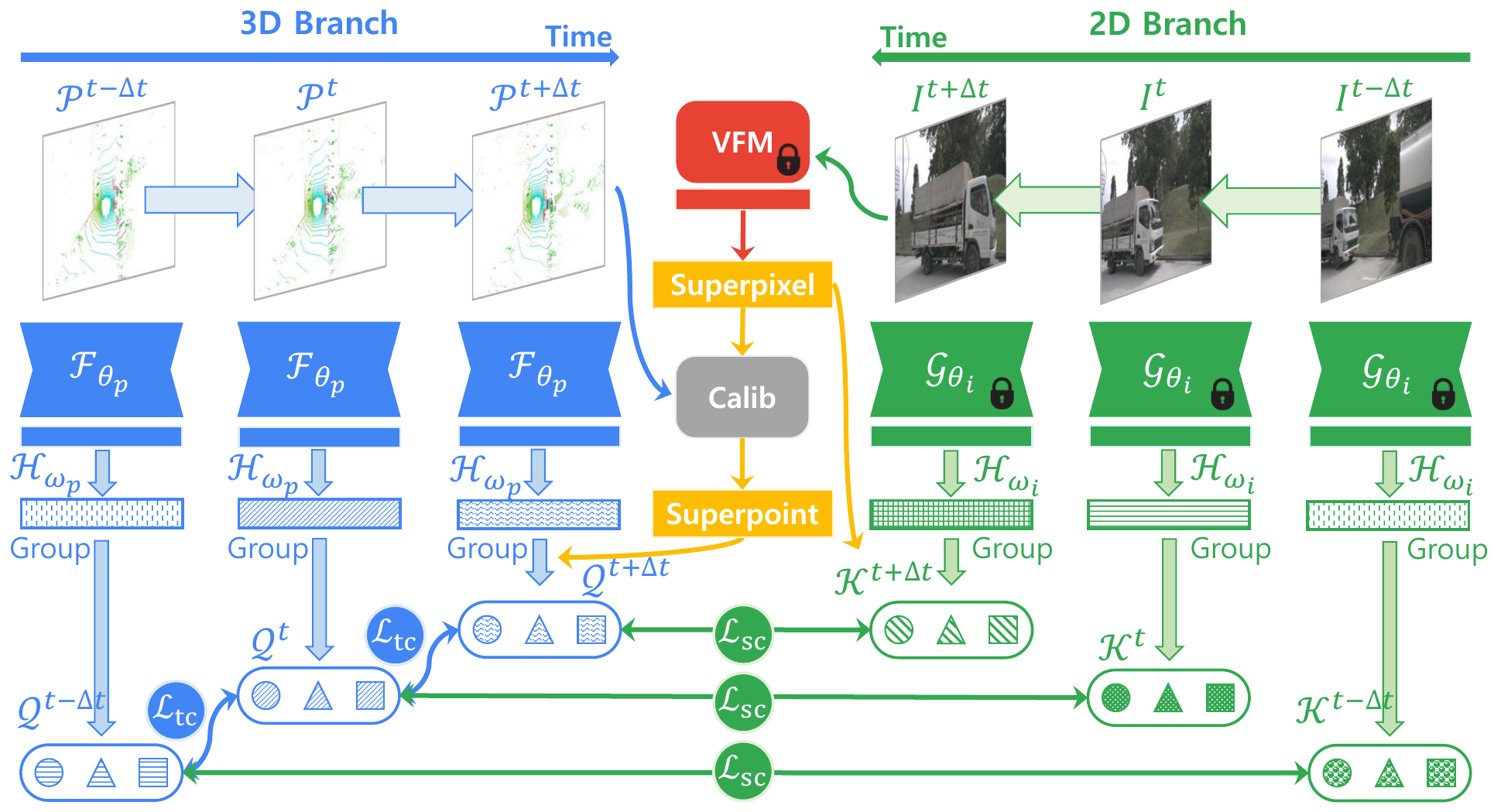}
    \end{center}
    \vspace{-0.4cm}
    \caption{\textbf{Flow-based contrastive learning (FCL) pipeline.} FCL takes multiple LiDAR-camera pairs from consecutive scans as input. Based on temporally aligned semantic superpixel and superpoints, two contrastive learning objectives are formulated: 1) spatial contrastive learning between each LiDAR-camera pair ($\mathcal{L}_{\text{sc}}$), and 2) temporal contrastive learning among consecutive LiDAR point clouds across scenes ($\mathcal{L}_{\text{tc}}$).}
    \label{fig:framework}
    \vspace{-0.2cm}
\end{figure}

\subsection{FCL: Flow-Based Contrastive Learning}
\label{sec:spatial_temporal}

\noindent\textbf{Motivation.}
LiDAR point clouds are acquired sequentially, embedding rich dynamic scene information across consecutive timestamps. Prior works \cite{slidr,st-slidr,seal} primarily focused on single LiDAR scans, overlooking the consistency of moving objects across scenes. To address these limitations, we propose flow-based contrastive learning (FCL) across sequential LiDAR scenes to encourage spatiotemporal consistency.

\noindent\textbf{Spatial Contrastive Learning.}
Our framework, depicted in \cref{fig:framework}, takes three LiDAR-camera pairs from different timestamps within a suitable time window as input, \ie, $\{(\mathcal{P}^{t}, \mathcal{I}^{t}), (\mathcal{P}^{t + \Delta t}, \mathcal{I}^{t + \Delta t}), (\mathcal{P}^{t - \Delta t}, \mathcal{I}^{t - \Delta t})\}$, where timestamp $t$ denotes the current scene and $\Delta t$ is the timespan. Following previous works \cite{slidr,seal}, we first distill knowledge from the 2D network into the 3D network for each scene separately. Taking $(\mathcal{P}^{t}, \mathcal{I}^{t})$ as an example, $\mathcal{P}^{t}$ and $\mathcal{I}^{t}$ are fed into the 3D and 2D networks to extract per-point and image features. The output features are then grouped via average pooling based on superpoints $\mathcal{Y}^{t}$ and superpixels $\mathcal{X}^{t}$ to obtain superpoint features $\mathbf{Q}^{t}$ and superpixel features $\mathbf{K}^{t}$. A spatial contrastive loss is formulated to constrain 3D representation via pretrained 2D prior knowledge. This process is formulated as follows:
\begin{equation}
    \mathcal{L}_{\text{sc}} = - \frac{1}{V} \sum_{i = 1}^{V} \log \left[ \frac{e^{(<\mathbf{q}_{i}, \mathbf{k}_{i}> / \tau)}}{\sum_{j \neq i}e^{(<\mathbf{q}_{i}, \mathbf{k}_{j}> / \tau)} + e^{(<\mathbf{q}_{i}, \mathbf{k}_{i}> / \tau)}} \right]~,
    \label{equ:spatial}
\end{equation}
where $\tau > 0$ is a temperature that controls the smoothness of distillation.

\noindent\textbf{Flow-Based Contrastive Learning.}
The spatial contrastive learning objective between images and point clouds, as depicted in \cref{equ:spatial}, fails to ensure that moving objects share similar attributes across different scenes. To maintain consistency across scenes, a temporal consistency loss is introduced among superpoint features across different scenes. For the point clouds $\mathcal{P}^{t}$ and $\mathcal{P}^{t + \Delta t}$, the corresponding superpoint features $\mathbf{Q}^{t}$ and $\mathbf{Q}^{t + \Delta t}$ are obtained via their superpoints. The temporal contrastive loss operates on $\mathbf{Q}^{t}$ and $\mathbf{Q}^{t + \Delta t}$:
\begin{equation}
    \mathcal{L}_{\text{tc}}^{t \leftarrow t + \Delta t} = - \frac{1}{V} \sum_{i = 1}^{V} \log \left[ \frac{e^{(<\mathbf{q}_{i}^{t}, \mathbf{q}_{i}^{t + \Delta t}> / \tau)}}{\sum_{j \neq i}e^{(<\mathbf{q}_{i}^{t}, \mathbf{q}_{j}^{t + \Delta t}> / \tau)} + e^{(<\mathbf{q}_{i}^{t}, \mathbf{q}_{i}^{t + \Delta t}> / \tau)}} \right]~.
    \label{equ:temporal}
\end{equation}
The same function is also applied between $\mathbf{Q}^{t}$ and $\mathbf{Q}^{t - \Delta t}$. This approach enables point features at time $t$ to extract more context-aware information across scenes.

\section{Experiments}
\label{sec:experiments}

\subsection{Settings}

\noindent\textbf{Data.} 
We follow the seminar works SLidR \cite{slidr} and Seal \cite{seal} when preparing the datasets. A total of eleven datasets are used in our experiments, including $^1$\textit{nuScenes} \cite{Panoptic-nuScenes}, $^2$\textit{SemanticKITTI} \cite{SemanticKITTI}, $^3$\textit{Waymo Open} \cite{sun2020waymoOpen}, $^4$\textit{ScribbleKITTI} \cite{unal2022scribbleKITTI}, $^5$\textit{RELLIS-3D} \cite{jiang2021rellis3D}, $^6$\textit{SemanticPOSS} \cite{pan2020semanticPOSS}, $^7$\textit{SemanticSTF} \cite{xiao2023semanticSTF}, $^8$\textit{SynLiDAR} \cite{xiao2022synLiDAR}, $^9$\textit{DAPS-3D} \cite{klokov2023daps3D}, $^{10}$\textit{Synth4D} \cite{saltori2020synth4D}, and $^{11}$\textit{Robo3D} \cite{kong2023robo3D}. Due to space limits, kindly refer to the Appendix and \cite{slidr,seal} for additional details about these datasets.

\begin{table}[t]
\centering
\caption{\textbf{Comparisons of state-of-the-art pretraining methods} pretrained on \textit{nuScenes}~\cite{Panoptic-nuScenes} and fine-tuned on \textit{SemanticKITTI} \cite{SemanticKITTI} and \textit{Waymo Open} \cite{sun2020waymoOpen} with specified 
data portions, respectively. All methods use MinkUNet \cite{choy2019minkowski} as the 3D semantic segmentation backbone. \textbf{LP} denotes linear probing with a frozen backbone. All scores are given in percentage (\%). Best scores in each configuration are shaded with colors.}
\vspace{-0.2cm}
\label{tab:benchmark}
\scalebox{0.76}{
\begin{tabular}{r|p{1.76cm}<{\raggedleft}|p{1.44cm}<{\raggedleft}|p{1.05cm}<{\centering}p{1.05cm}<{\centering}p{1.05cm}<{\centering}p{1.05cm}<{\centering}p{1.05cm}<{\centering}p{1.05cm}<{\centering}|p{1.28cm}<{\centering}|p{1.28cm}<{\centering}}
\toprule
\multirow{2}{*}{\textbf{Method}} & \multirow{2}{*}{\textbf{Venue}} & \multirow{2}{*}{\textbf{Distill}} & \multicolumn{6}{c}{\textbf{nuScenes}} \vline & \textbf{KITTI} & \textbf{Waymo}
\\
& & & {\textbf{LP}} & {$\mathbf{1\%}$} & {$\mathbf{5\%}$} & {$\mathbf{10\%}$} & {$\mathbf{25\%}$} & {\textbf{Full}} & {$\mathbf{1\%}$} & {$\mathbf{1\%}$}
\\\midrule\midrule
\textcolor{gray}{Random} & \textcolor{gray}{-} & \textcolor{gray}{-} & \textcolor{gray}{$8.10$} & \textcolor{gray}{$30.30$} & \textcolor{gray}{$47.84$} & \textcolor{gray}{$56.15$} & \textcolor{gray}{$65.48$} & \textcolor{gray}{$74.66$} & \textcolor{gray}{$39.50$} & \textcolor{gray}{$39.41$}
\\\midrule
PointContrast~\cite{pointcontrast} & ECCV'20 & None~\textcolor{sf_gray}{$\circ$} & $21.90$ & $32.50$ & - & - & - & - & $41.10$ & -
\\
DepthContrast~\cite{depthcontrast} & ICCV'21 & None~\textcolor{sf_gray}{$\circ$} & \cellcolor{sf_gray!18}$22.10$ & $31.70$ & - & - & - & - & \cellcolor{sf_gray!18}$41.50$ & -
\\
ALSO~\cite{boulch2023also} & CVPR'23 & None~\textcolor{sf_gray}{$\circ$} & - & $37.70$ & - & $59.40$ & - & $72.00$ & - & -
\\
BEVContrast~\cite{sautier2023bevcontrast} & 3DV'24 & None~\textcolor{sf_gray}{$\circ$} & - & \cellcolor{sf_gray!18}$38.30$ & - & \cellcolor{sf_gray!18}$59.60$ & - & \cellcolor{sf_gray!18}$72.30$ & - & -
\\\midrule
PPKT~\cite{ppkt} & arXiv'21 & ResNet~\textcolor{sf_blue}{$\circ$} & $35.90$ & $37.80$ & $53.74$ & $60.25$ & $67.14$ & $74.52$ & $44.00$ & $47.60$
\\
SLidR~\cite{slidr} & CVPR'22 & ResNet~\textcolor{sf_blue}{$\circ$} & $38.80$ & $38.30$ & $52.49$ & $59.84$ & $66.91$ & $74.79$ & $44.60$ & $47.12$
\\
ST-SLidR~\cite{st-slidr} & CVPR'23 & ResNet~\textcolor{sf_blue}{$\circ$} & $40.48$ & $40.75$ & $54.69$ & $60.75$ & $67.70$ & $75.14$ & $44.72$ & $44.93$
\\
TriCC~\cite{pang2023tricc} & CVPR'23 & ResNet~\textcolor{sf_blue}{$\circ$} & $38.00$ & $41.20$ & $54.10$ & $60.40$ & $67.60$ & $75.60$ & $45.90$ & -
\\
Seal~\cite{seal} & NeurIPS'23 & ResNet~\textcolor{sf_blue}{$\circ$} & \cellcolor{sf_blue!13}$44.95$ & \cellcolor{sf_blue!13}$45.84$ & $55.64$ & \cellcolor{sf_blue!13}$62.97$ & $68.41$ & $75.60$ & $46.63$ & \cellcolor{sf_blue!13}$49.34$
\\
HVDistill~\cite{zhang2024hvdistill} & IJCV'24 & ResNet~\textcolor{sf_blue}{$\circ$} & $39.50$ & $42.70$ & \cellcolor{sf_blue!13}$56.60$ & $62.90$ & \cellcolor{sf_blue!13}$69.30$ & \cellcolor{sf_blue!13}$76.60$ & \cellcolor{sf_blue!13}$49.70$ & -
\\\midrule
PPKT~\cite{ppkt} & arXiv'21 & ViT-S~\textcolor{sf_red}{$\circ$} & $38.60$ & $40.60$ & $52.06$ & $59.99$ & $65.76$ & $73.97$ & $43.25$ & $47.44$
\\
SLidR~\cite{slidr} & CVPR'22 & ViT-S~\textcolor{sf_red}{$\circ$} & $44.70$ & $41.16$ & $53.65$ & $61.47$ & $66.71$ & $74.20$ & $44.67$ & $47.57$
\\
Seal~\cite{seal} & NeurIPS'23 & ViT-S~\textcolor{sf_red}{$\circ$} & $45.16$ & $44.27$ & $55.13$ & $62.46$ & $67.64$ & $75.58$ & $46.51$ & $48.67$
\\
\textbf{SuperFlow} & \textbf{Ours} & ViT-S~\textcolor{sf_red}{$\bullet$} & \cellcolor{sf_red!10}$46.44$ & \cellcolor{sf_red!10}$47.81$ & \cellcolor{sf_red!10}$59.44$ & \cellcolor{sf_red!10}$64.47$ & \cellcolor{sf_red!10}$69.20$ & \cellcolor{sf_red!10}$76.54$ & \cellcolor{sf_red!10}$47.97$ & \cellcolor{sf_red!10}$49.94$
\\\midrule
PPKT~\cite{ppkt} & arXiv'21 & ViT-B~\textcolor{sf_yellow}{$\circ$} & $39.95$ & $40.91$ & $53.21$ & $60.87$ & $66.22$ & $74.07$ & $44.09$ & $47.57$
\\
SLidR~\cite{slidr} & CVPR'22 & ViT-B~\textcolor{sf_yellow}{$\circ$} & $45.35$ & $41.64$ & $55.83$ & $62.68$ & $67.61$ & $74.98$ & $45.50$ & $48.32$ 
\\
Seal~\cite{seal} & NeurIPS'23 & ViT-B~\textcolor{sf_yellow}{$\circ$} & $46.59$ & $45.98$ & $57.15$ & $62.79$ & $68.18$ & $75.41$ & $47.24$ & $48.91$
\\
\textbf{SuperFlow} & \textbf{Ours} & ViT-B~\textcolor{sf_yellow}{$\bullet$} & \cellcolor{sf_yellow!12}$47.66$ & \cellcolor{sf_yellow!12}$48.09$ & \cellcolor{sf_yellow!12}$59.66$ & \cellcolor{sf_yellow!12}$64.52$ & \cellcolor{sf_yellow!12}$69.79$ & \cellcolor{sf_yellow!12}$76.57$ & \cellcolor{sf_yellow!12}$48.40$ & \cellcolor{sf_yellow!12}$50.20$
\\\midrule
PPKT~\cite{ppkt} & arXiv'21 & ViT-L~\textcolor{sf_green}{$\circ$} & $41.57$ & $42.05$ & $55.75$ & $61.26$ & $66.88$ & $74.33$ & $45.87$ & $47.82$
\\
SLidR~\cite{slidr} & CVPR'22 & ViT-L~\textcolor{sf_green}{$\circ$} & $45.70$ & $42.77$ & $57.45$ & $63.20$ & $68.13$ & $75.51$ & $47.01$ & $48.60$
\\
Seal~\cite{seal} & NeurIPS'23 & ViT-L~\textcolor{sf_green}{$\circ$} & $46.81$ & $46.27$ & $58.14$ & $63.27$ & $68.67$ & $75.66$ & $47.55$ & $50.02$
\\
\textbf{SuperFlow} & \textbf{Ours} & ViT-L~\textcolor{sf_green}{$\bullet$} & \cellcolor{sf_green!12}$48.01$ & \cellcolor{sf_green!12}$49.95$ & \cellcolor{sf_green!12}$60.72$ & \cellcolor{sf_green!12}$65.09$ & \cellcolor{sf_green!12}$70.01$ & \cellcolor{sf_green!12}$77.19$ & \cellcolor{sf_green!12}$49.07$ & \cellcolor{sf_green!12}$50.67$
\\\bottomrule
\end{tabular}
}
\vspace{-0.2cm}
\end{table}

\noindent\textbf{Implementation Details.}
SuperFlow is implemented using the MMDetection3D \cite{mmdet3d} and OpenPCSeg \cite{pcseg2023} codebases. Consistent with prior works \cite{slidr,seal}, we employ MinkUNet \cite{choy2019minkowski} as the 3D backbone and DINOv2 \cite{oquab2023dinov2} (with ViT backbones \cite{dosovitskiy2020vit}) as the 2D backbone, distilling from three variants: small (S), base (B), and large (L). Following Seal \cite{seal}, OpenSeeD \cite{zhang2023openSeeD} is used to generate semantic superpixels. The framework is pretrained end-to-end on $600$ scenes from \textit{nuScenes} \cite{Panoptic-nuScenes}, then linear probed and fine-tuned on \textit{nuScenes} \cite{Panoptic-nuScenes} according to the data splits in SLidR \cite{slidr}. The domain generalization study adheres to the same configurations as Seal \cite{seal} for the other ten datasets. Both the baselines and SuperFlow are pretrained using eight GPUs for $50$ epochs, while linear probing and downstream fine-tuning experiments use four GPUs for $100$ epochs, all utilizing the AdamW optimizer \cite{AdamW} and OneCycle scheduler \cite{OneCycle}. Due to space limits, kindly refer to the Appendix for additional implementation details.

\noindent\textbf{Evaluation Protocols.} Following conventions, we report the Intersection-over-Union (IoU) on each semantic class and mean IoU (mIoU) over all classes for downstream tasks. For 3D robustness evaluations, we follow Robo3D \cite{kong2023robo3D} and report the mean Corruption Error (mCE) and mean Resilience Rate (mRR).

\begin{table}[t]
\centering
\caption{\textbf{Domain generalization study} of different pretraining methods pretrained on the \textit{nuScenes}~\cite{Panoptic-nuScenes} dataset and fine-tuned on other \textit{seven} heterogeneous 3D semantic segmentation datasets with specified data portions, respectively. All scores are given in percentage (\%). Best scores in each configuration are shaded with colors.}
\vspace{-0.2cm}
\label{tab:multiple_datasets}
\scalebox{0.76}{
\begin{tabular}{r|  p{0.9cm}<{\centering}p{0.9cm}<{\centering}|p{0.9cm}<{\centering}p{0.9cm}<{\centering}|p{0.9cm}<{\centering}p{0.9cm}<{\centering}|p{0.9cm}<{\centering}p{0.9cm}<{\centering}|p{0.9cm}<{\centering}p{0.9cm}<{\centering}|p{0.9cm}<{\centering}p{0.9cm}<{\centering}|p{0.9cm}<{\centering}p{0.9cm}<{\centering}}
\toprule
\multirow{2}{*}{\textbf{Method}} & \multicolumn{2}{c}{\textbf{ScriKITTI}} \vline & \multicolumn{2}{c}{\textbf{Rellis-3D}} \vline & \multicolumn{2}{c}{\textbf{SemPOSS}} \vline & \multicolumn{2}{c}{\textbf{SemSTF}} \vline & \multicolumn{2}{c}{\textbf{SynLiDAR}} \vline & \multicolumn{2}{c}{\textbf{DAPS-3D}} \vline & \multicolumn{2}{c}{\textbf{Synth4D}} 
\\
& {$\mathbf{1\%}$} & {$\mathbf{10\%}$} & {$\mathbf{1\%}$} & {$\mathbf{10\%}$} & {\textbf{Half}} & {\textbf{Full}} & {\textbf{Half}} & {\textbf{Full}} & {$\mathbf{1\%}$} & {$\mathbf{10\%}$} & {\textbf{Half}} & {\textbf{Full}} & {$\mathbf{1\%}$} & {$\mathbf{10\%}$}
\\\midrule\midrule
\textcolor{gray}{Random} & \textcolor{gray}{$23.81$} & \textcolor{gray}{$47.60$} & \textcolor{gray}{$38.46$} & \textcolor{gray}{$53.60$} & \textcolor{gray}{$46.26$} & \textcolor{gray}{$54.12$} & \textcolor{gray}{$48.03$} & \textcolor{gray}{$48.15$} & \textcolor{gray}{$19.89$} & \textcolor{gray}{$44.74$} & \textcolor{gray}{$74.32$} &  \textcolor{gray}{$79.38$} & \textcolor{gray}{$20.22$} & \textcolor{gray}{$66.87$}
\\\midrule
PPKT~\cite{ppkt} & $36.50$ & $51.67$ & $49.71$ & $54.33$ & $50.18$ & $56.00$ & $50.92$ & $54.69$ & $37.57$ & $46.48$ & $78.90$ & $84.00$ & $61.10$ & $62.41$
\\
SLidR~\cite{slidr} & $39.60$ & $50.45$ & $49.75$ & $54.57$ & $51.56$ & $55.36$ & $52.01$ & $54.35$ & $42.05$ & $47.84$ & $81.00$  & $85.40$ & $63.10$ & $62.67$
\\
Seal~\cite{seal} & $40.64$ & $52.77$ & $51.09$ & $55.03$ & $53.26$ & $56.89$ & $53.46$ & $55.36$ & $43.58$ & $49.26$ & $81.88$ & $85.90$ & $64.50$ & $66.96$
\\
\textbf{SuperFlow} & \cellcolor{sf_blue!13}$42.70$ & \cellcolor{sf_blue!13}$54.00$ & \cellcolor{sf_blue!13}$52.83$ & \cellcolor{sf_blue!13}$55.71$ & \cellcolor{sf_blue!13}$54.41$ & \cellcolor{sf_blue!13}$57.33$ & \cellcolor{sf_blue!13}$54.72$ & \cellcolor{sf_blue!13}$56.57$ & \cellcolor{sf_blue!13}$44.85$ & \cellcolor{sf_blue!13}$51.38$ & \cellcolor{sf_blue!13}$82.43$ & \cellcolor{sf_blue!13}$86.21$ & \cellcolor{sf_blue!13}$65.31$ & \cellcolor{sf_blue!13}$69.43$
\\\bottomrule
\end{tabular}
}
\vspace{-0.2cm}
\end{table}
\begin{table}[t]
\centering
\caption{\textbf{Out-of-distribution 3D robustness study} of state-of-the-art pretraining methods under corruption and sensor failure scenarios in the \textit{nuScenes-C} dataset from the \textit{Robo3D} benchmark \cite{kong2023robo3D}. \textbf{Full} denotes fine-tuning with full labels. \textbf{LP} denotes linear probing with a frozen backbone. All mCE ($\downarrow$), mRR ($\uparrow$), and mIoU ($\uparrow$) scores are given in percentage (\%). Best scores in each configuration are shaded with colors.}
\vspace{-0.2cm}
\label{tab:robo3d}
\scalebox{0.76}{
\begin{tabular}{c|r|r|p{1.1cm}<{\centering}|p{1.1cm}<{\centering}|p{0.93cm}<{\centering}p{0.93cm}<{\centering}p{0.93cm}<{\centering}p{0.93cm}<{\centering}p{0.93cm}<{\centering}p{0.93cm}<{\centering}p{0.93cm}<{\centering}p{0.93cm}<{\centering}|p{0.96cm}<{\centering}}
\toprule
\textbf{\#} & \textbf{Initial} & \textbf{Backbone} & \textbf{mCE} & \textbf{mRR} & Fog & Rain & Snow & Blur & Beam & Cross & Echo & Sensor & \textbf{Avg}
\\\midrule\midrule
\multirow{12}{*}{\rotatebox[origin=c]{90}{\textbf{Full}}} & \textcolor{gray}{Random} & MinkU-18~\textcolor{sf_gray}{$\circ$} & \textcolor{gray}{$115.61$} & \textcolor{gray}{$70.85$} & \textcolor{gray}{$53.90$} & \textcolor{gray}{$71.10$} & \textcolor{gray}{$48.22$} & \textcolor{gray}{$51.85$} & \textcolor{gray}{$62.21$} & \textcolor{gray}{$37.73$} & \textcolor{gray}{$57.47$} & \textcolor{gray}{$38.97$} & \textcolor{gray}{$52.68$}
\\
& \textbf{SuperFlow} & MinkU-18~\textcolor{sf_gray}{$\bullet$} & \cellcolor{sf_gray!18}$109.00$ & \cellcolor{sf_gray!18}$75.66$ & \cellcolor{sf_gray!18}$54.95$ & \cellcolor{sf_gray!18}$72.79$ & \cellcolor{sf_gray!18}$49.56$ & \cellcolor{sf_gray!18}$57.68$ & \cellcolor{sf_gray!18}$62.82$ & \cellcolor{sf_gray!18}$42.45$ & \cellcolor{sf_gray!18}$59.61$ & \cellcolor{sf_gray!18}$41.77$ & \cellcolor{sf_gray!18}$55.21$
\\\cmidrule{2-14}
& \textcolor{gray}{Random} & MinkU-34~\textcolor{sf_blue}{$\circ$} & \textcolor{gray}{$112.20$} & \textcolor{gray}{$72.57$} & \textcolor{gray}{$62.96$} & \textcolor{gray}{$70.65$} & \textcolor{gray}{$55.48$} & \textcolor{gray}{$51.71$} & \textcolor{gray}{$62.01$} & \textcolor{gray}{$31.56$} & \textcolor{gray}{$59.64$} & \textcolor{gray}{$39.41$} & \textcolor{gray}{$54.18$}
\\
& PPKT \cite{ppkt} & MinkU-34~\textcolor{sf_blue}{$\circ$} & $105.64$ & $75.87$ & $64.01$ & $72.18$ & $59.08$ & $57.17$ & $63.88$ & $36.34$ & $60.59$ & $39.57$ & $56.60$
\\
& SLidR \cite{slidr} & MinkU-34~\textcolor{sf_blue}{$\circ$} & $106.08$ & $75.99$ & $65.41$ & $72.31$ & $56.01$ & $56.07$ & $62.87$ & $41.94$ & $61.16$ & $38.90$ & $56.83$
\\
& Seal \cite{seal} & MinkU-34~\textcolor{sf_blue}{$\circ$} & $92.63$ & $83.08$ & \cellcolor{sf_blue!13}$72.66$ & $74.31$ & \cellcolor{sf_blue!13}$66.22$ & \cellcolor{sf_blue!13}$66.14$ & $65.96$ & $57.44$ & \cellcolor{sf_blue!13}$59.87$ & $39.85$ & $62.81$
\\
& \textbf{SuperFlow} & MinkU-34~\textcolor{sf_blue}{$\bullet$} & \cellcolor{sf_blue!13}$91.67$ & \cellcolor{sf_blue!13}$83.17$ & $70.32$ & \cellcolor{sf_blue!13}$75.77$ & $65.41$ & $61.05$ & \cellcolor{sf_blue!13}$68.09$ & \cellcolor{sf_blue!13}$60.02$ & $58.36$ & \cellcolor{sf_blue!13}$50.41$ & \cellcolor{sf_blue!13}$63.68$
\\\cmidrule{2-14}
& \textcolor{gray}{Random} & MinkU-50~\textcolor{sf_red}{$\circ$} & \textcolor{gray}{$113.76$} & \textcolor{gray}{$72.81$} & \textcolor{gray}{$49.95$} & \textcolor{gray}{$71.16$} & \textcolor{gray}{$45.36$} & \textcolor{gray}{$55.55$} & \textcolor{gray}{$62.84$} & \textcolor{gray}{$36.94$} & \textcolor{gray}{$59.12$} & \textcolor{gray}{$43.15$} & \textcolor{gray}{$53.01$}
\\
& \textbf{SuperFlow} & MinkU-50~\textcolor{sf_red}{$\bullet$} & \cellcolor{sf_red!10}$107.35$ & \cellcolor{sf_red!10}$74.02$ & \cellcolor{sf_red!10}$54.36$ & \cellcolor{sf_red!10}$73.08$ & \cellcolor{sf_red!10}$50.07$ & \cellcolor{sf_red!10}$56.92$ & \cellcolor{sf_red!10}$64.05$ & \cellcolor{sf_red!10}$38.10$ & \cellcolor{sf_red!10}$62.02$ & \cellcolor{sf_red!10}$47.02$ & \cellcolor{sf_red!10}$55.70$
\\\cmidrule{2-14}
& \textcolor{gray}{Random} & MinkU-101~\textcolor{sf_yellow}{$\circ$} & \textcolor{gray}{$109.10$} & \textcolor{gray}{$74.07$} & \textcolor{gray}{$50.45$} & \textcolor{gray}{$73.02$} & \textcolor{gray}{$48.85$} & \textcolor{gray}{$58.48$} & \textcolor{gray}{$64.18$} & \textcolor{gray}{$43.86$} & \textcolor{gray}{$59.82$} & \textcolor{gray}{$41.47$} & \textcolor{gray}{$55.02$}
\\
& \textbf{SuperFlow} & MinkU-101~\textcolor{sf_yellow}{$\bullet$} & \cellcolor{sf_yellow!12}$96.44$ & \cellcolor{sf_yellow!12}$78.57$ & \cellcolor{sf_yellow!12}$56.92$ & \cellcolor{sf_yellow!12}$76.29$ & \cellcolor{sf_yellow!12}$54.70$ & \cellcolor{sf_yellow!12}$59.35$ & \cellcolor{sf_yellow!12}$71.89$ & \cellcolor{sf_yellow!12}$55.13$ & \cellcolor{sf_yellow!12}$60.27$ & \cellcolor{sf_yellow!12}$51.60$ & \cellcolor{sf_yellow!12}$60.77$
\\\midrule
\multirow{4}{*}{\rotatebox[origin=c]{90}{\textbf{LP}}}
& PPKT \cite{ppkt} & MinkU-34~\textcolor{sf_green}{$\circ$} & $183.44$ & \cellcolor{sf_green!12}$78.15$ & $30.65$ & $35.42$ & $28.12$ & $29.21$ & $32.82$ & $19.52$ & $28.01$ & $20.71$ & $28.06$
\\
& SLidR \cite{slidr} & MinkU-34~\textcolor{sf_green}{$\circ$} & $179.38$ & $77.18$ & $34.88$ & $38.09$ & $32.64$ & $26.44$ & $33.73$ & $20.81$ & $31.54$ & $21.44$ & $29.95$
\\
& Seal \cite{seal} & MinkU-34~\textcolor{sf_green}{$\circ$} & $166.18$ & $75.38$ & $37.33$ & $42.77$ & $29.93$ & $37.73$ & $40.32$ & $20.31$ & $37.73$ & $24.94$ & $33.88$
\\
& \textbf{SuperFlow} & MinkU-34~\textcolor{sf_green}{$\bullet$} & \cellcolor{sf_green!12}$161.78$ & $75.52$ & \cellcolor{sf_green!12}$37.59$ & \cellcolor{sf_green!12}$43.42$ & \cellcolor{sf_green!12}$37.60$ & \cellcolor{sf_green!12}$39.57$ & \cellcolor{sf_green!12}$41.40$ & \cellcolor{sf_green!12}$23.64$ & \cellcolor{sf_green!12}$38.03$ & \cellcolor{sf_green!12}$26.69$ & \cellcolor{sf_green!12}$35.99$
\\\bottomrule
\end{tabular}
}
\vspace{-0.2cm}
\end{table}

\subsection{Comparative Study}

\noindent\textbf{Linear Probing.}
We start by investigating the pretraining quality via linear probing. For this setup, we initialize the 3D backbone $\mathcal{F}_{\theta_{p}}$ with pretrained parameters and fine-tune only the added-on segmentation head. As shown in \cref{tab:benchmark}, SuperFlow consistently outperforms state-of-the-art methods under diverse configurations. We attribute this to the use of temporal consistency learning, which captures the structurally rich temporal cues across consecutive scenes and enhances the semantic representation learning of the 3D backbone. We also observe improved performance with larger 2D networks (\ie, from ViT-S to ViT-L), revealing a promising direction of achieving higher quality 3D pretraining.

\noindent\textbf{Downstream Fine-Tuning.}
It is known that data representation learning can mitigate the need for large-scale human annotations. Our study systematically compares SuperFlow with prior works on three popular datasets, including \textit{nuScenes} \cite{Panoptic-nuScenes}, \textit{SemanticKITTI} \cite{SemanticKITTI}, and \textit{Waymo Open} \cite{sun2020waymoOpen}, under limited annotations for few-shot fine-tuning. From \cref{tab:benchmark}, we observe that SuperFlow achieves promising performance gains among three datasets across all fine-tuning tasks. We also use the pretrained 3D backbone as initialization for the fully-supervised learning study on \textit{nuScenes} \cite{Panoptic-nuScenes}. As can be seen from \cref{tab:benchmark}, models pretrained via representation learning consistently outperform the random initialization counterparts, highlighting the efficacy of conducting data pretraining. We also find that distillations from larger 2D networks show consistent improvements.

\noindent\textbf{Cross-Domain Generalization.}
To verify the strong generalizability of SuperFlow, we conduct a comprehensive study using seven diverse LiDAR datasets and show results in \cref{tab:multiple_datasets}. It is worth noting that these datasets are collected under different acquisition and annotation conditions, including adverse weather, weak annotations, synthetic collection, and dynamic objects. For all fourteen domain generalization fine-tuning tasks, SuperFlow exhibits superior performance over the prior arts \cite{ppkt,slidr,seal}. This study strongly verifies the effectiveness of the proposed flow-based contrastive learning for image-to-LiDAR data representation.

\noindent\textbf{Out-of-Distribution Robustness.}
The robustness of 3D perception models against unprecedented conditions directly correlates with the model's applicability to real-world applications \cite{kong2023robodepth,xie2024robobev,hao2024mapbench,li2024place3d}. We compare our SuperFlow with prior models in the \textit{nuScenes-C} dataset from the \textit{Robo3D} benchmark \cite{kong2023robo3D} and show results in \cref{tab:robo3d}. We observe that models pretrained using SuperFlow exhibit improved robustness over the random initialization counterparts. Besides, we find that 3D networks with different capacities often pose diverse robustness.

\begin{table}[t]
\begin{minipage}{0.45\textwidth}
    \centering
    \captionof{table}{\textbf{Ablation study of SuperFlow} using different \# of sweeps. All methods use ViT-B \cite{oquab2023dinov2} for distillation. All scores are given in percentage (\%). Baseline results are shaded with colors.}
    \label{tab:sweeps}
    \vspace{-0.2cm}
    \scalebox{0.76}{\begin{tabular}{r|p{1.1cm}<{\centering}p{1.1cm}<{\centering}|p{1.28cm}<{\centering}|p{1.28cm}<{\centering}}
    \toprule
    \multirow{2}{*}{\textbf{Backbone}} & \multicolumn{2}{c}{\textbf{nuScenes}} \vline & \textbf{KITTI} & \textbf{Waymo}
    \\
    & \textbf{LP} & $\mathbf{1\%}$ & $\mathbf{1\%}$ & $\mathbf{1\%}$
    \\\midrule\midrule
    $1\times$ Sweeps~\textcolor{sf_blue}{$\circ$} & $47.41$ & $47.52$ & $48.14$ & $49.31$
    \\
    $2\times$ Sweeps~\textcolor{sf_blue}{$\bullet$} & \cellcolor{sf_blue!13}$47.66$ & \cellcolor{sf_blue!13}$48.09$ & \cellcolor{sf_blue!13}$48.40$ & \cellcolor{sf_blue!13}$50.20$
    \\
    $5\times$ Sweeps~\textcolor{sf_blue}{$\circ$} & $47.23$ & $48.00$ & $47.94$ & $49.14$
    \\
    $7\times$ Sweeps~\textcolor{sf_blue}{$\circ$} & $46.03$ & $47.98$ & $46.83$ & $47.97$
    \\\bottomrule
    \end{tabular}}
\end{minipage}~~~
\begin{minipage}{0.51\textwidth}
    \centering
    \captionof{table}{\textbf{Ablation study of SuperFlow} on network capacity (\# params) of 3D backbones. All methods use ViT-B \cite{oquab2023dinov2} for distillation. All scores are given in percentage (\%). Baseline results are shaded with colors.}
    \label{tab:network_capacity}
    \vspace{-0.2cm}
    \scalebox{0.76}{\begin{tabular}{r|c|p{1.1cm}<{\centering}p{1.1cm}<{\centering}|p{1.28cm}<{\centering}|p{1.28cm}<{\centering}}
    \toprule
    \multirow{2}{*}{\textbf{Backbone}} & \multirow{2}{*}{\textbf{Layer}} & \multicolumn{2}{c}{\textbf{nuScenes}} \vline & \textbf{KITTI} & \textbf{Waymo}
    \\
    & & \textbf{LP} & $\mathbf{1\%}$ & $\mathbf{1\%}$  & $\mathbf{1\%}$
    \\\midrule\midrule
    MinkUNet~\textcolor{sf_blue}{$\circ$} & $18$ & $47.20$ & $47.70$ & $48.04$ & $49.24$
    \\
    MinkUNet~\textcolor{sf_blue}{$\bullet$} & $34$ & \cellcolor{sf_blue!13}$47.66$ & \cellcolor{sf_blue!13}$48.09$ & \cellcolor{sf_blue!13}$48.40$ & \cellcolor{sf_blue!13}$50.20$
    \\
    MinkUNet~\textcolor{sf_blue}{$\circ$} & $50$ & $54.11$ & $52.86$ & $49.22$ & $51.20$
    \\
    MinkUNet~\textcolor{sf_blue}{$\circ$} & $101$ & $52.56$ & $51.19$ & $48.51$ & $50.01$
    \\\bottomrule
    \end{tabular}}
\end{minipage}
\end{table}

\begin{figure}[!ht]
    \begin{center}
    \includegraphics[width=\textwidth]{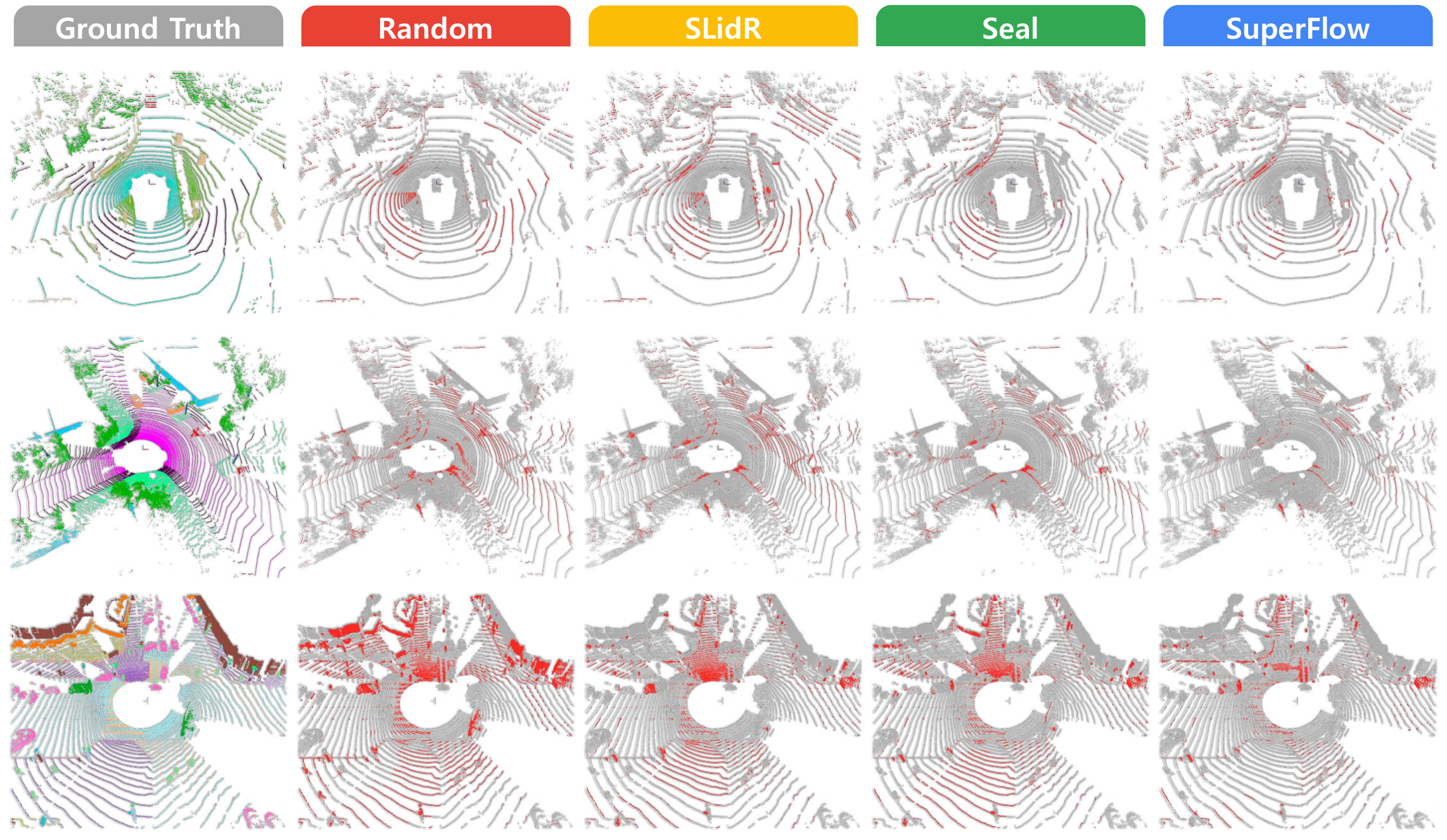}
    \end{center}
    \vspace{-0.5cm}
    \caption{\textbf{Qualitative assessments} of state-of-the-art pretraining methods pretrained on \textit{nuScenes} \cite{Panoptic-nuScenes} and fine-tuned on \textit{nuScenes} \cite{Panoptic-nuScenes}, \textit{SemanticKITTI} \cite{SemanticKITTI}, and \textit{Waymo Open} \cite{sun2020waymoOpen}, with $1\%$ annotations. The error maps show the \textcolor{gray}{correct} and \textcolor{sf_red}{incorrect} predictions in \textcolor{gray}{gray} and \textcolor{sf_red}{red}, respectively. Best viewed in colors and zoomed-in for details.}
    \label{fig:qualitative}
    \vspace{-0.2cm}
\end{figure}

\noindent\textbf{Quantitative Assessments.} We visualize the prediction results fine-tuned on nuScenes \cite{Panoptic-nuScenes}, SemanticKITTI \cite{SemanticKITTI} and Waymo Open \cite{sun2020waymoOpen}, compared with random initialization, SLiDR \cite{slidr}, and Seal \cite{seal}. As shown in \cref{fig:qualitative}, Superflow performs well, especially on backgrounds, \ie, ``road'' and ``sidewalk'' in complex scenarios.

\subsection{Ablation Study}

In this section, we are tailored to understand the efficacy of each design in our SuperFlow framework. Unless otherwise specified, we adopt MinkUNet-34 \cite{choy2019minkowski} and ViT-B \cite{oquab2023dinov2} as the 3D and 2D backbones, respectively, throughout this study.

\noindent\textbf{3D Network Capacity.}
Existing 3D backbones are relatively small in scale compared to their 2D counterparts. We study the scale of the 3D network and the results are shown in \cref{tab:network_capacity}. We observe improved performance as the network capacity scales up, except for MinkUNet-101 \cite{choy2019minkowski}. We conjecture that this is due to the fact that models with limited parameters are less effective in capturing patterns during representation learning, and, conversely, models with a large set of trainable parameters tend to be difficult to converge.

\begin{table}[t]
\begin{minipage}{0.505\textwidth}
    \centering
    \captionof{table}{\textbf{Ablation study of each component in SuperFlow}. All variants use a MinkUNet-34 \cite{choy2019minkowski} as the 3D backbone and ViT-B \cite{oquab2023dinov2} for distillation. \textbf{VC}: View consistency. \textbf{D2S}: Dense-to-sparse regularization. \textbf{FCL}: Flow-based contrastive learning. All scores are given in percentage (\%).}
    \vspace{-0.2cm}
    \label{tab:ablation}
    \scalebox{0.76}{
    \begin{tabular}{c|p{0.75cm}<{\centering}p{0.75cm}<{\centering}p{0.75cm}<{\centering}|p{1cm}<{\centering}p{1cm}<{\centering}|p{1.27cm}<{\centering}|p{1.27cm}<{\centering}}
    \toprule
    \multirow{2}{*}{\textbf{\#}} &
    \multirow{2}{*}{\textbf{VC}} & \multirow{2}{*}{\textbf{D2S}} & \multirow{2}{*}{\textbf{FCL}} & \multicolumn{2}{c}{\textbf{nuScenes}} \vline & \textbf{KITTI} & \textbf{Waymo}
    \\
    & & & & {\textbf{LP}} & {$\mathbf{1\%}$} &{$\mathbf{1\%}$} & {$\mathbf{1\%}$}
    \\\midrule\midrule
    \textcolor{gray}{-} & \multicolumn{3}{c|}{\textcolor{gray}{Random}} & \textcolor{gray}{$8.10$} & \textcolor{gray}{$30.30$} & \textcolor{gray}{$39.50$} & \textcolor{gray}{$39.41$}
    \\\midrule
    (a) & \textcolor{sf_red}{\xmark} & \textcolor{sf_red}{\xmark} & \textcolor{sf_red}{\xmark} & $44.65$ & $44.47$ & $46.65$ & $47.77$
    \\
    (b) & \textcolor{sf_blue}{\cmark} & \textcolor{sf_red}{\xmark} & \textcolor{sf_red}{\xmark} & $45.57$ & $45.21$ & $46.87$ & $48.01$
    \\
    (c) & \textcolor{sf_blue}{\cmark} & \textcolor{sf_blue}{\cmark} & \textcolor{sf_red}{\xmark} & $46.17$ & $46.91$ & $47.26$ & $49.01$
    \\
    (d) & \textcolor{sf_blue}{\cmark} & \textcolor{sf_red}{\xmark} & \textcolor{sf_blue}{\cmark} & $47.24$ & $47.67$ & $48.21$ & $49.80$
    \\
    (e) & \textcolor{sf_blue}{\cmark} & \textcolor{sf_blue}{\cmark} & \textcolor{sf_blue}{\cmark} & \cellcolor{sf_blue!13}$47.66$ & \cellcolor{sf_blue!13}$48.09$ & \cellcolor{sf_blue!13}$48.40$ & \cellcolor{sf_blue!13}$50.20$
    \\\bottomrule
    \end{tabular}}
\end{minipage}~~~~
\begin{minipage}{0.455\textwidth}
    \centering
    \captionof{table}{\textbf{Ablation study on spatiotemporal consistency}. All variants use a MinkUNet-34 \cite{choy2019minkowski} as the 3D backbone and ViT-B \cite{oquab2023dinov2} for distillation. $\mathbf{0}$ denotes current timestamp. $\mathbf{0.5s}$ corresponds to a $20$Hz timespan. All scores are given in percentage (\%).}
    \vspace{-0.2cm}
    \label{tab:temporal}
    \scalebox{0.76}{\begin{tabular}{c|p{1cm}<{\centering}p{1cm}<{\centering}|p{1.27cm}<{\centering}|p{1.27cm}<{\centering}}
    \toprule
    \multirow{2}{*}{\textbf{Timespan}} & \multicolumn{2}{c}{\textbf{nuScenes}} \vline & \textbf{KITTI} & \textbf{Waymo}
    \\
    & \textbf{LP} & $\mathbf{1\%}$ & $\mathbf{1\%}$  & $\mathbf{1\%}$
    \\\midrule\midrule
    \textcolor{gray}{Single-Frame} & \textcolor{gray}{$46.17$} & \textcolor{gray}{$46.91$} & \textcolor{gray}{$47.26$} & \textcolor{gray}{$49.01$}
    \\\midrule
    $0, -0.5s$ & $46.39$ & $47.08$ & $47.99$ & $49.78$
    \\
    $-0.5s, 0, +0.5s$ & \cellcolor{sf_blue!13}$47.66$ & \cellcolor{sf_blue!13}$48.09$ & \cellcolor{sf_blue!13}$48.40$ & \cellcolor{sf_blue!13}$50.20$
    \\
    $-1.0s, 0, +1.0s$ & $47.60$ & $47.99$ & $48.43$ & $50.18$
    \\
    $-1.5s, 0, +1.5s$ & $46.43$ & $48.27$ & $48.34$ & $49.93$
    \\
    $-2.0s, 0, +2.0s$ & $46.20$ & $48.49$ & $48.18$ & $50.01$
    \\\bottomrule
    \end{tabular}}
\end{minipage}
\vspace{-0.3cm}
\end{table}

\noindent\textbf{Representation Density.}
The consistency regularization between sparse and dense point clouds encourages useful representation learning. To analyze the degree of regularization, we investigate various point cloud densities and show the results in \cref{tab:sweeps}. We observe that a suitable point cloud density can improve the model's ability to feature representation. When the density of point clouds is too dense, the motion of objects is obvious in the scene. However, we generate superpoints of the dense points based on superpixels captured at the time of sparse points. The displacement difference of dynamic objects makes the projection misalignment. A trade-off selection would be two or three sweeps.

\noindent\textbf{Temporal Consistency.}
The ability to capture semantically coherent temporal cues is crucial in our SuperFlow framework. In \cref{equ:temporal}, we operate temporal contrastive learning on superpoints features across scenes. As shown in \cref{tab:temporal}, we observe that temporal contrastive learning achieves better results compared to single-frame methods. We also compare the impact of frames used to capture temporal cues. When we use $3$ frames, it acquires more context-aware information than $2$ frames and achieves better results. Finally, we study the impact of the timespan between frames. The performance will drop with a longer timespan. We conjecture that scenes with short timespans have more consistency, while long timespans tend to have more uncertain factors.

\noindent\textbf{Component Analysis.}
In \cref{tab:ablation}, we analyze each component in the SuperFlow framework, including view consistency, dense-to-sparse regularization, and flow-based contrastive learning. The baseline is SLiDR \cite{slidr} with VFMs-based superpixels. View consistency brings slight improvements among the popular datasets with a few annotations. D2S distills dense features into sparse features and it brings about $1\%$ mIoU gains. FCL extracts temporal cues via temporal contrastive learning and it significantly leads to about $2.0\%$ mIoU gains.

\begin{figure}[t]
    \centering
    \begin{subfigure}[b]{0.306\textwidth}
        \centering
        \includegraphics[width=\textwidth]{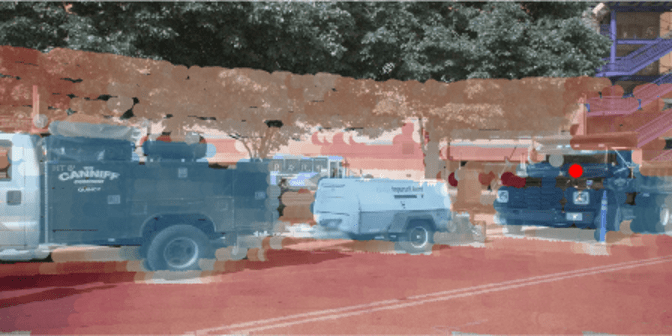}
        \caption{``car'' (3D)}
        \label{fig:car_3d}
    \end{subfigure}
    ~~
    \begin{subfigure}[b]{0.306\textwidth}
        \centering
        \includegraphics[width=\textwidth]{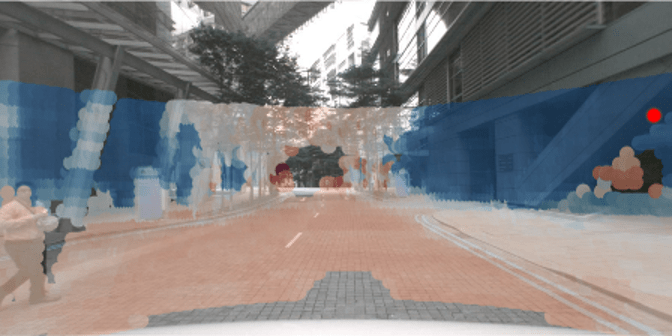}
        \caption{``manmade'' (3D)}
        \label{fig:manmade_3d}
    \end{subfigure}
    ~~
    \begin{subfigure}[b]{0.306\textwidth}
        \centering
        \includegraphics[width=\textwidth]{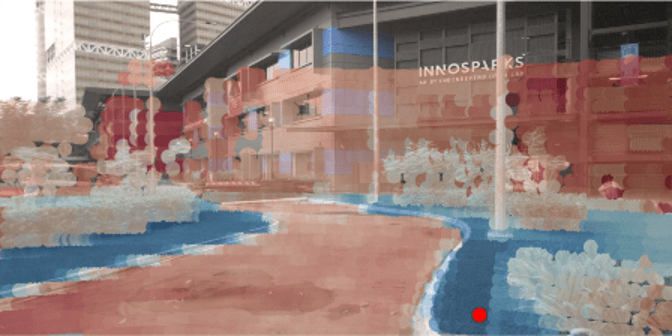}
        \caption{``sidewalk'' (3D)}
        \label{fig:sidewalk_3d}
    \end{subfigure}
    \begin{subfigure}[b]{0.306\textwidth}
        \centering
        \includegraphics[width=\textwidth]{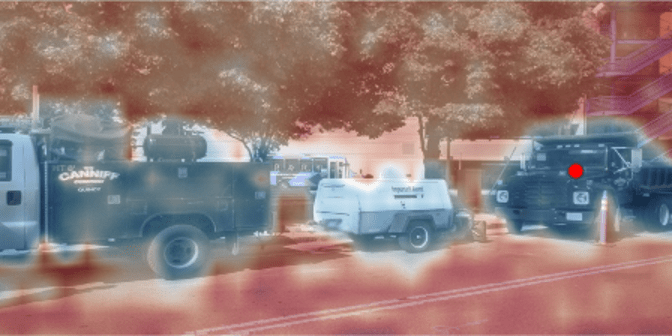}
        \caption{``car'' (2D)}
        \label{fig:car_2d}
    \end{subfigure}
    ~~
    \begin{subfigure}[b]{0.306\textwidth}
        \centering
        \includegraphics[width=\textwidth]{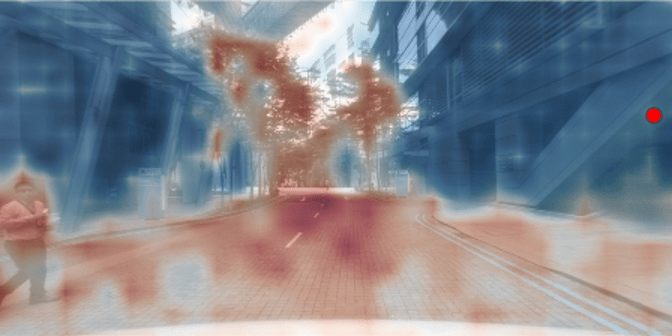}
        \caption{``manmade'' (2D)}
        \label{fig:manmade_2d}
    \end{subfigure}
    ~~
    \begin{subfigure}[b]{0.306\textwidth}
        \centering
        \includegraphics[width=\textwidth]{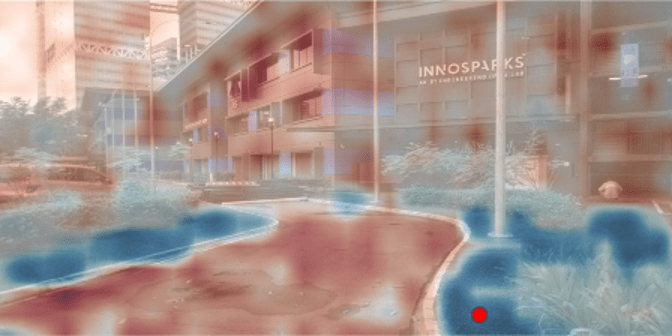}
        \caption{``sidewalk'' (2D)}
        \label{fig:sidewalk_2d}
    \end{subfigure}
    \begin{subfigure}[b]{0.306\textwidth}
        \centering
        \includegraphics[width=\textwidth]{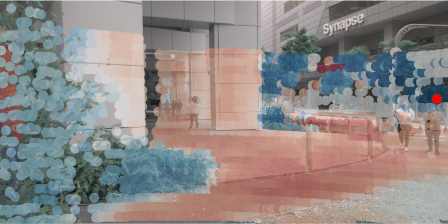}
        \caption{``vegetation'' (3D)}
        \label{fig:vegetation_3d}
    \end{subfigure}
    ~~
    \begin{subfigure}[b]{0.306\textwidth}
        \centering
        \includegraphics[width=\textwidth]{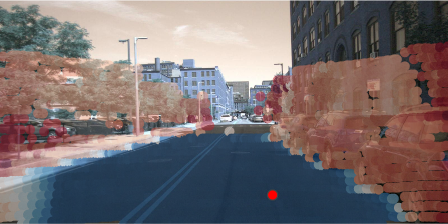}
        \caption{``driveable surface'' (3D)}
        \label{fig:driveable_3d}
    \end{subfigure}
    ~~
    \begin{subfigure}[b]{0.306\textwidth}
        \centering
        \includegraphics[width=\textwidth]{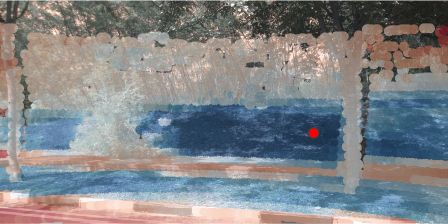}
        \caption{``terrain'' (3D)}
        \label{fig:terrain_3d}
    \end{subfigure}
    \begin{subfigure}[b]{0.306\textwidth}
        \centering
        \includegraphics[width=\textwidth]{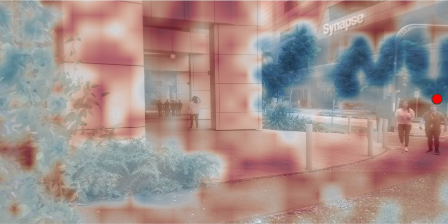}
        \caption{``vegetation'' (2D)}
        \label{fig:vegetation_2d}
    \end{subfigure}
    ~~
    \begin{subfigure}[b]{0.306\textwidth}
        \centering
        \includegraphics[width=\textwidth]{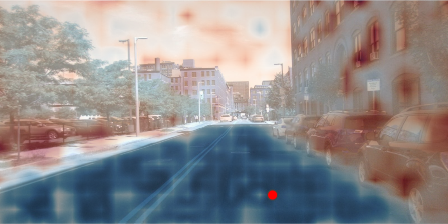}
        \caption{``driveable surface'' (2D)}
        \label{fig:driveable_2d}
    \end{subfigure}
    ~~
    \begin{subfigure}[b]{0.306\textwidth}
        \centering
        \includegraphics[width=\textwidth]{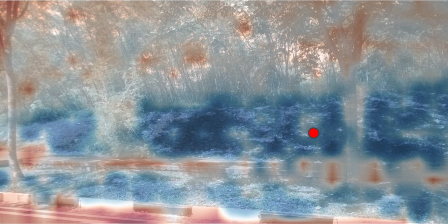}
        \caption{``terrain'' (2D)}
        \label{fig:terrain_2d}
    \end{subfigure}
    \caption{\textbf{Cosine similarity} between features of a query point (\textcolor{red}{red dot}) and: 1) features of other points projected in the image (the 1st and 3rd rows); and 2) features of an image with the same scene (the 2nd and 4th rows). The color goes from \textcolor{red}{red} to \textcolor{blue}{blue} denoting low and high similarity scores, respectively. Best viewed in color.}
    \label{fig:similarity}
    \vspace{-0.2cm}
\end{figure}

\noindent\textbf{Visual Inspections.}
Similarity maps presented in \cref{fig:similarity} denote the segmentation ability of our pretrained model. The query points include ``car'', ``manmade'', ``sidewalk'', ``vegetation'', ``driveable surface'', and ``terrain''. SuperFlows shows strong semantic discriminative ability without fine-tuning. We conjecture that it comes from three aspects: 1) View consistent superpixels enable the network to learn semantic representation; 2) Dense-to-sparse regularization enhances the network to learn varying density features; 3) Temporal contrastive learning extracts semantic cues across scenes.

\section{Conclusion}
\label{sec:conclusion}

In this work, we presented \textbf{SuperFlow} to tackle the challenging 3D data representation learning. Motivated by the sequential nature of LiDAR acquisitions, we proposed three novel designs to better encourage spatiotemporal consistency, encompassing view consistency alignment, dense-to-sparse regularization, and flow-based contrastive learning. Extensive experiments across 11 diverse LiDAR datasets showed that SuperFlow consistently outperforms prior approaches in linear probing, downstream fine-tuning, and robustness probing. Our study on scaling up 2D and 3D network capacities reveals insightful findings. We hope this work could shed light on future designs of powerful 3D foundation models.

{\noindent\textbf{Acknowledgements.}
This work was supported by the Scientific and Technological Innovation 2030 - ``New Generation Artificial Intelligence'' Major Project (No. 2021ZD0112200), the Joint Funds of the National Natural Science Foundation of China (No. U21B2044), the Key Research and Development Program of Jiangsu Province (No. BE2023016-3), and the Talent Research Start-up Foundation of Nanjing University of Posts and Telecommunications (No. NY223172). This work was also supported by the Ministry of Education, Singapore, under its MOE AcRF Tier 2 (MOET2EP20221- 0012), NTU NAP, and under the RIE2020 Industry Alignment Fund – Industry Collaboration Projects (IAF-ICP) Funding Initiative, as well as cash and in-kind contribution from the industry partner(s).
}

\section*{Appendix}
\begin{itemize}

    \item \textcolor{eccvblue}{\textbf{6. \nameref{sec:supp_implement}}} \dotfill \pageref{sec:supp_implement}
    \begin{itemize}
        \item \textcolor{eccvblue}{6.1 \nameref{subsec:datasets}} \dotfill \pageref{subsec:datasets}
        \item \textcolor{eccvblue}{6.2 \nameref{subsec:training}} \dotfill \pageref{subsec:training}
        \item \textcolor{eccvblue}{6.3 \nameref{subsec:evaluation}} \dotfill \pageref{subsec:evaluation}
    \end{itemize}

    \item \textcolor{eccvblue}{\textbf{7. \nameref{sec:supp_quantitative}}} \dotfill \pageref{sec:supp_quantitative}
    \begin{itemize}
        \item \textcolor{eccvblue}{7.1 \nameref{subsec:linear_probe}} \dotfill \pageref{subsec:linear_probe}
        \item \textcolor{eccvblue}{7.2 \nameref{subsec:fine_tune}} \dotfill \pageref{subsec:fine_tune}
    \end{itemize}

    \item \textcolor{eccvblue}{\textbf{8. \nameref{sec:supp_qualitative}}} \dotfill \pageref{sec:supp_qualitative}
    \begin{itemize}
        \item \textcolor{eccvblue}{8.1 \nameref{subsec:lidarseg}} \dotfill \pageref{subsec:lidarseg}
        \item \textcolor{eccvblue}{8.2 \nameref{subsec:cosine_similarity}} \dotfill \pageref{subsec:cosine_similarity}
    \end{itemize}

    \item \textcolor{eccvblue}{\textbf{9. \nameref{sec:supp_limitation}}} \dotfill \pageref{sec:supp_limitation}
    \begin{itemize}
        \item \textcolor{eccvblue}{9.1 \nameref{subsec:potential_limitation}} \dotfill \pageref{subsec:potential_limitation}
        \item \textcolor{eccvblue}{9.2 \nameref{subsec:potential_societal_impact}} \dotfill \pageref{subsec:potential_societal_impact}
    \end{itemize}

    \item \textcolor{eccvblue}{\textbf{10. \nameref{sec:supp_acknowledge}}} \dotfill \pageref{sec:supp_acknowledge}
    \begin{itemize}
        \item \textcolor{eccvblue}{10.1 \nameref{subsec:acknowledge_codebase}} \dotfill \pageref{subsec:acknowledge_codebase}
        \item \textcolor{eccvblue}{10.2 \nameref{subsec:acknowledge_datasets}} \dotfill \pageref{subsec:acknowledge_datasets}
        \item \textcolor{eccvblue}{10.3 \nameref{subsec:acknowledge_implements}} \dotfill \pageref{subsec:acknowledge_implements}
    \end{itemize}

\end{itemize}

\section{Additional Implementation Detail}
\label{sec:supp_implement}

In this section, we elaborate on additional details regarding the datasets, hyperparameters, and training/evaluation configuration.

\subsection{Datasets}
\label{subsec:datasets}

\begin{table}[t]
\caption{\textbf{Summary of datasets used in this work.} Our study encompasses a total of 10 datasets in the linear probing and downstream generalization experiments, including $^1$\textit{nuScenes} \cite{Panoptic-nuScenes}, $^2$\textit{SemanticKITTI} \cite{SemanticKITTI}, $^3$\textit{Waymo Open} \cite{sun2020waymoOpen}, $^4$\textit{ScribbleKITTI} \cite{unal2022scribbleKITTI}, $^5$\textit{RELLIS-3D} \cite{jiang2021rellis3D}, $^6$\textit{SemanticPOSS} \cite{pan2020semanticPOSS}, $^7$\textit{SemanticSTF} \cite{xiao2023semanticSTF}, $^8$\textit{SynLiDAR} \cite{xiao2022synLiDAR}, $^9$\textit{DAPS-3D} \cite{klokov2023daps3D}, $^{10}$\textit{Synth4D} \cite{saltori2020synth4D}, and $^{11}$\textit{nuScenes-C} \cite{kong2023robo3D}. Images adopted from the original papers.}
\vspace{-0.2cm}
\centering
\resizebox{\textwidth}{!}{
\begin{tabular}{c|c|c|c|c}
    \toprule
    nuScenes & SemanticKITTI & Waymo Open & ScribbleKITTI & RELLIS-3D
    \\\midrule
    \begin{minipage}[b]{0.2\columnwidth}\centering\raisebox{-.3\height}{\includegraphics[width=\linewidth]{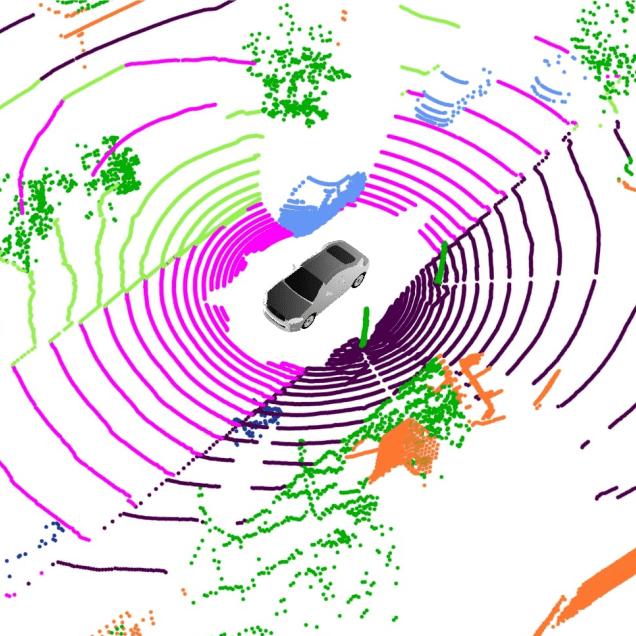}}\end{minipage} & 
    \begin{minipage}[b]{0.2\columnwidth}\centering\raisebox{-.3\height}{\includegraphics[width=\linewidth]{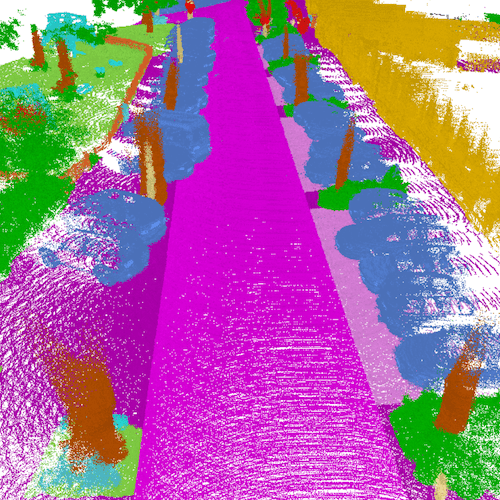}}\end{minipage} & 
    \begin{minipage}[b]{0.2\columnwidth}\centering\raisebox{-.3\height}{\includegraphics[width=\linewidth]{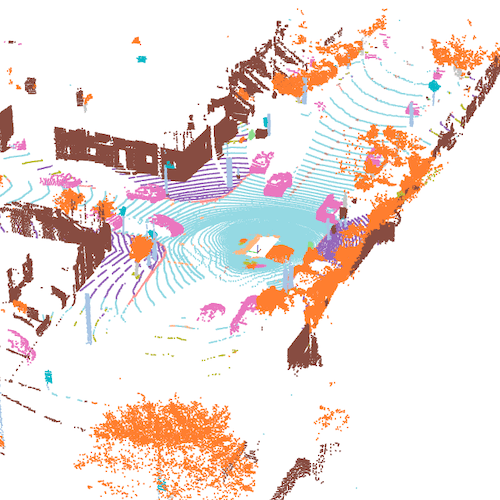}}\end{minipage} & 
    \begin{minipage}[b]{0.2\columnwidth}\centering\raisebox{-.3\height}{\includegraphics[width=\linewidth]{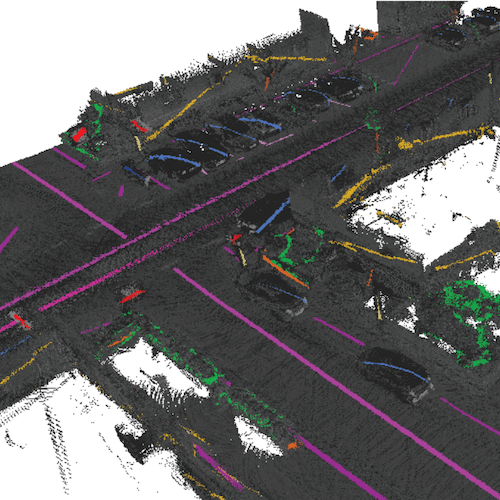}}\end{minipage} & 
    \begin{minipage}[b]{0.2\columnwidth}\centering\raisebox{-.3\height}{\includegraphics[width=\linewidth]{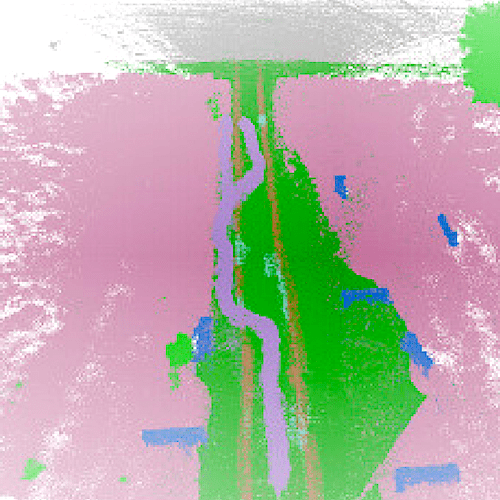}}\end{minipage}
    \\\midrule
    SemanticPOSS & SemanticSTF & SynLiDAR & DAPS-3D & Synth4D
    \\\midrule
    \begin{minipage}[b]{0.2\columnwidth}\centering\raisebox{-.3\height}{\includegraphics[width=\linewidth]{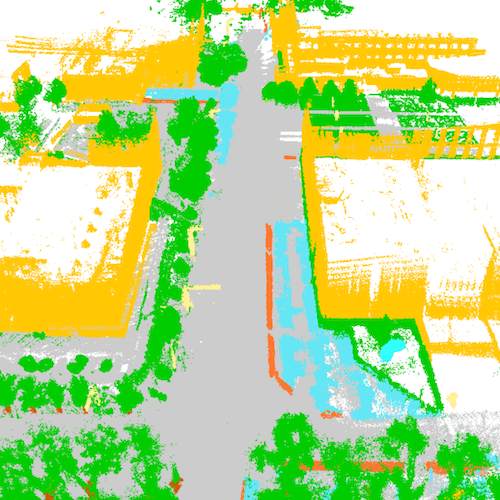}}\end{minipage} & \begin{minipage}[b]{0.2\columnwidth}\centering\raisebox{-.3\height}{\includegraphics[width=\linewidth]{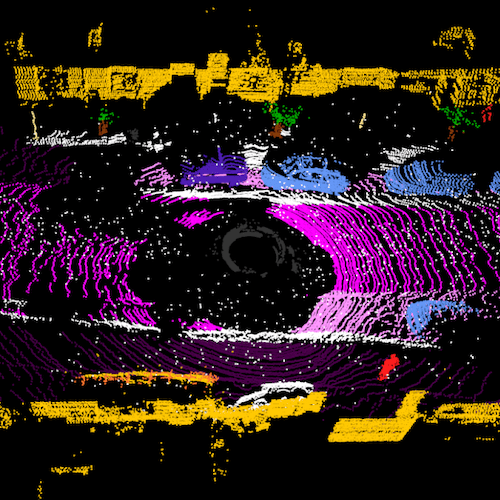}}\end{minipage}
    & 
    \begin{minipage}[b]{0.2\columnwidth}\centering\raisebox{-.3\height}{\includegraphics[width=\linewidth]{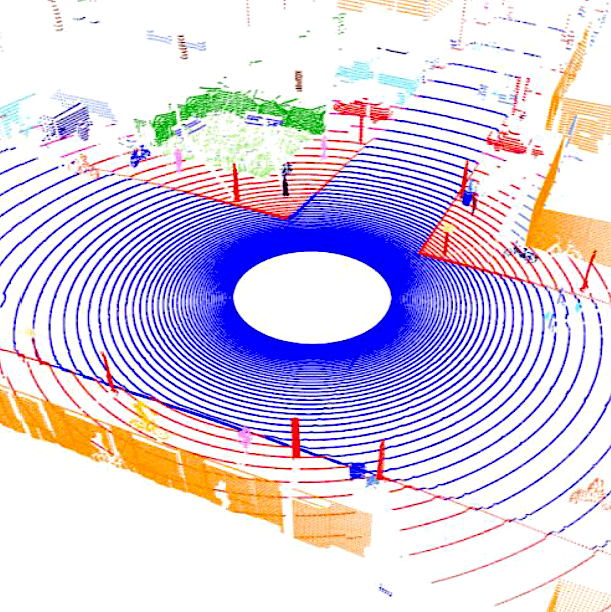}}\end{minipage} & 
    \begin{minipage}[b]{0.2\columnwidth}\centering\raisebox{-.3\height}{\includegraphics[width=\linewidth]{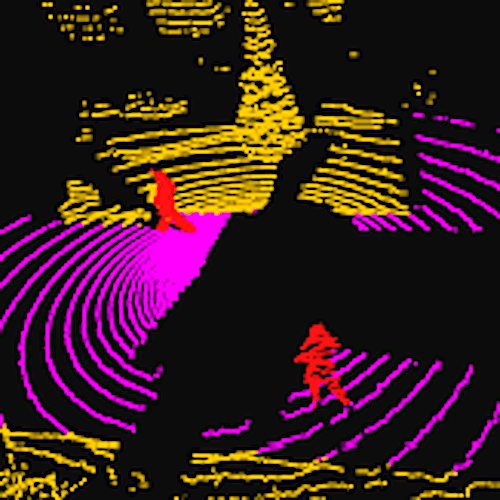}}\end{minipage} &
    \begin{minipage}[b]{0.2\columnwidth}\centering\raisebox{-.3\height}{\includegraphics[width=\linewidth]{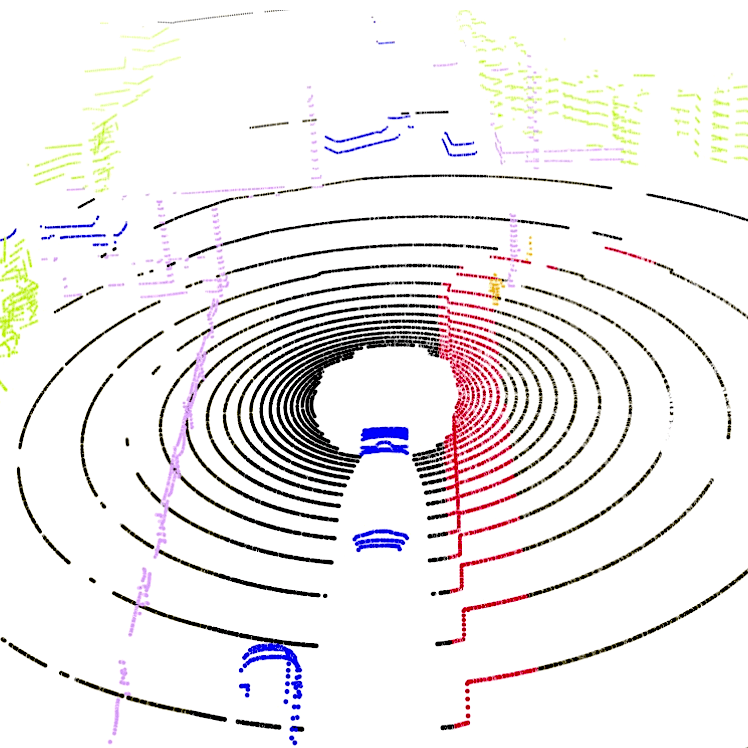}}\end{minipage}
    \\\bottomrule
\end{tabular}}
\label{tab:dataset_summary}
\vspace{-0.2cm}
\end{table}

\begin{table}[t]
\caption{\textbf{Examples of the out-of-distribution (OoD) scenarios.} Our study encompasses a total of 8 common OoD scenarios in the 3D robustness evaluation experiments, including $^1$\textit{fog}, $^2$\textit{wet ground}, $^3$\textit{snow}, $^4$\textit{motion blur}, $^5$\textit{beam missing}, $^6$\textit{crosstalk}, $^7$\textit{incomplete echo}, and $^8$\textit{cross sensor}. Images adopted from the Robo3D \cite{kong2023robo3D} paper.}
\vspace{-0.2cm}
\centering
\scalebox{0.89}{
\begin{tabular}{p{3.232cm}<{\centering}|p{3.232cm}<{\centering}|p{3.232cm}<{\centering}|p{3.232cm}<{\centering}}
\toprule
    Fog & Wet Ground & Snow & Motion Blur
    \\\midrule
    \begin{minipage}[b]{0.25\columnwidth}\centering\raisebox{-.3\height}{\includegraphics[width=\linewidth]{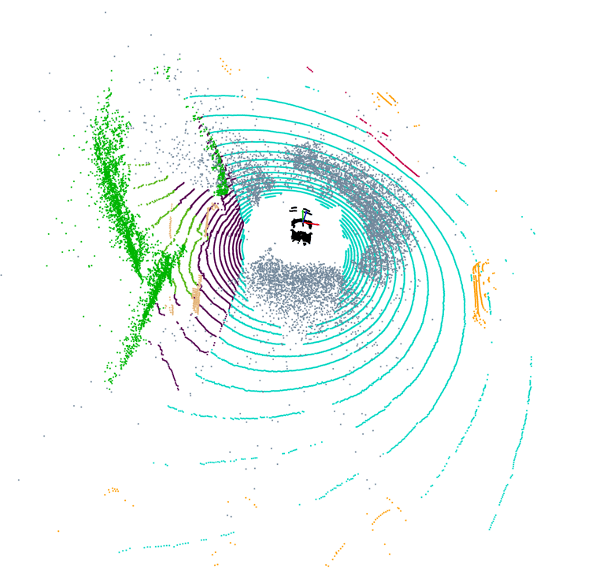}}\end{minipage} & 
    \begin{minipage}[b]{0.25\columnwidth}\centering\raisebox{-.3\height}{\includegraphics[width=\linewidth]{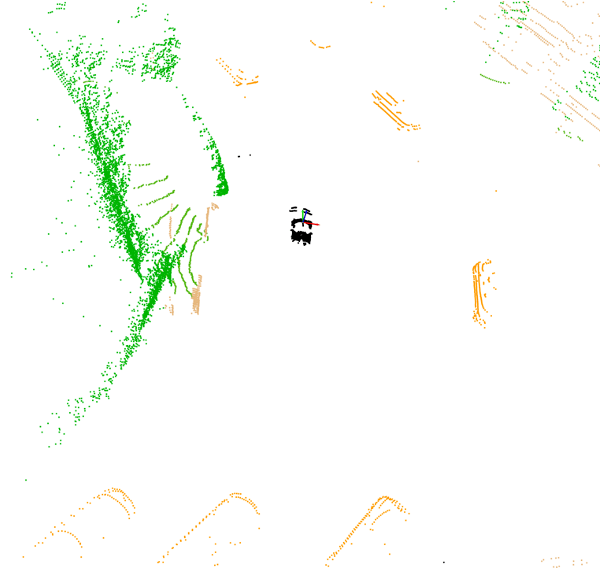}}\end{minipage} & 
    \begin{minipage}[b]{0.25\columnwidth}\centering\raisebox{-.3\height}{\includegraphics[width=\linewidth]{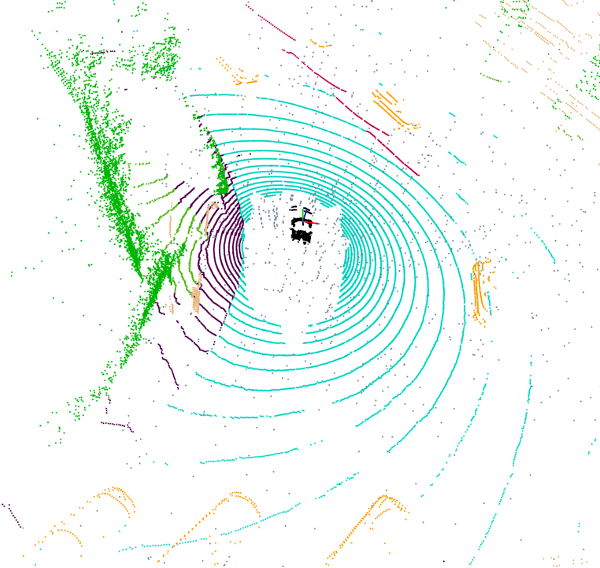}}\end{minipage} & 
    \begin{minipage}[b]{0.25\columnwidth}\centering\raisebox{-.3\height}{\includegraphics[width=\linewidth]{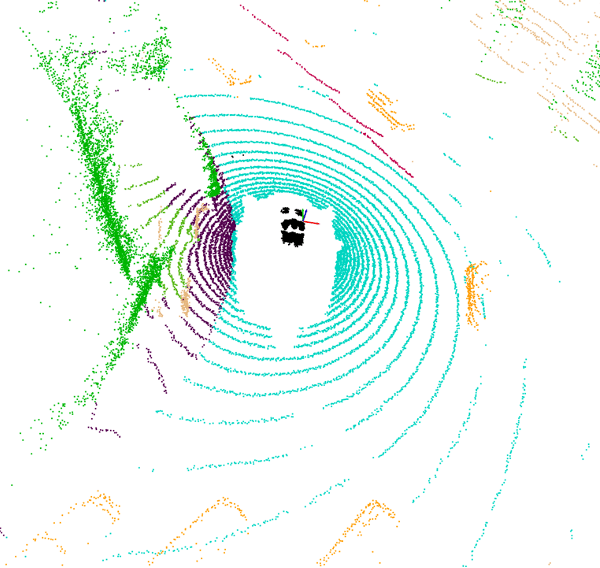}}\end{minipage}
    \\\midrule
    Beam Missing & Crosstalk & Incomplete Echo & Cross Sensor
    \\\midrule
    \begin{minipage}[b]{0.25\columnwidth}\centering\raisebox{-.3\height}{\includegraphics[width=\linewidth]{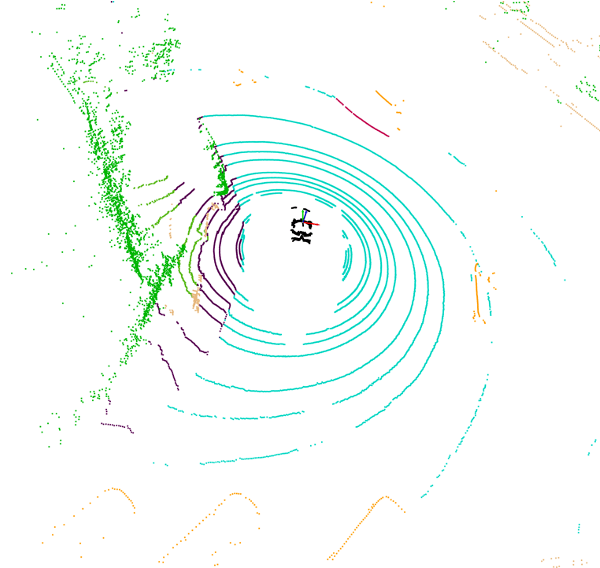}}\end{minipage} & \begin{minipage}[b]{0.25\columnwidth}\centering\raisebox{-.3\height}{\includegraphics[width=\linewidth]{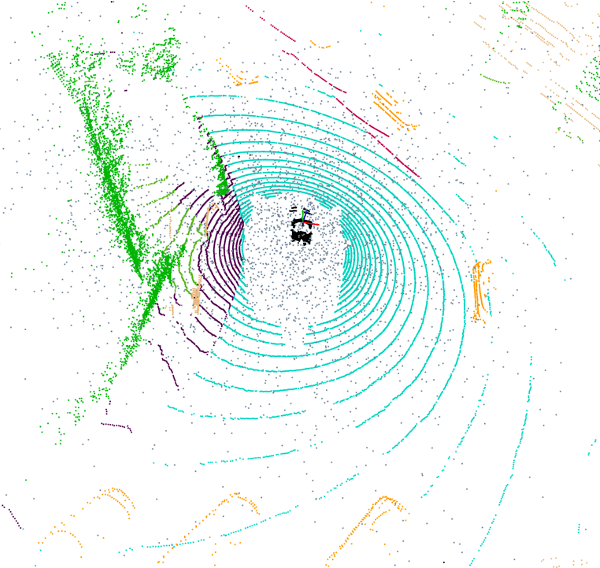}}\end{minipage}
    & 
    \begin{minipage}[b]{0.25\columnwidth}\centering\raisebox{-.3\height}{\includegraphics[width=\linewidth]{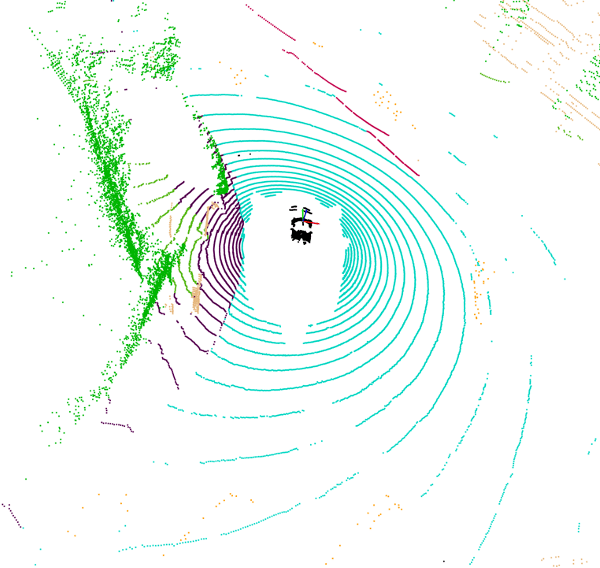}}\end{minipage} & 
    \begin{minipage}[b]{0.25\columnwidth}\centering\raisebox{-.3\height}{\includegraphics[width=\linewidth]{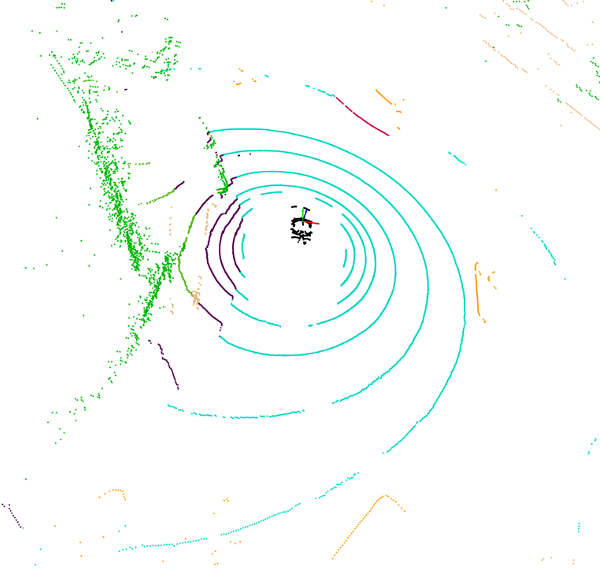}}\end{minipage}
    \\\bottomrule
\end{tabular}}
\label{tab:robo3d_dataset}
\vspace{-0.2cm}
\end{table}

\noindent\textbf{Pretraining.}
In this work, we pretrain the model on the \textit{nuScenes} \cite{Panoptic-nuScenes} dataset following the data split in SLidR \cite{slidr}. Specifically, $600$ scenes are used as the training set for model pretraining, which is a mini-train split of the whole $700$ training scenes. It includes both LiDAR point clouds and six camera image data, from labeled keyframe data to multiple unlabeled sweeps. We conduct spatiotemporal contrastive learning with keyframe data and dense-to-sparse regularization by combining multiple LiDAR sweeps to form dense points.

\noindent\textbf{Linear Probing.}
We train the 3D backbone network with the fixed pretrained backbone on the training set of \textit{nuScenes} \cite{Panoptic-nuScenes}, and evaluate the performance on the validation set. It consists of $700$ training scenes (for 29,130 samples) and 150 validation scenes (for 6,019 samples). Following the conventional setup, the evaluation results are calculated among $16$ merged semantic categories.

\noindent\textbf{Downstream Fine-Tuning.}
To validate the pretraining quality of each self-supervised learning approach, we conduct a comprehensive downstream fine-tuning experiment on the nuScenes \cite{Panoptic-nuScenes} dataset, with various configurations. Specifically, we train the 3D backbone network with the pretrained backbone using $1\%$, $5\%$, $10\%$, $25\%$, and $100\%$ annotated data, respectively, and evaluate the model's performance on the official validation set.

\noindent\textbf{Cross-Domain Fine-Tuning.}
In this work, we conduct a comprehensive cross-domain fine-tuning experiment on a total of 9 datasets. \cref{tab:dataset_summary} provides a summary of these datasets. Specifically, \textit{SemanticKITTI} \cite{SemanticKITTI} and \textit{Waymo Open} \cite{sun2020waymoOpen} contain large-scale LiDAR scans collected from real-world driving scenes, which are acquired by 64-beam LiDAR sensors. We construct the $1\%$ training sample set by sampling every $100$ frame from the whole training set. \textit{ScribbleKITTI} \cite{unal2022scribbleKITTI} shares the same scene with \textit{SemanticKITTI} \cite{SemanticKITTI} but are weakly annotated with line scribbles. The total percentage of valid annotated labels is $8.06\%$ compared to fully-supervised methods, while saving about $90\%$ annotation times. \textit{RELLIS-3D} \cite{jiang2021rellis3D} is a multimodal dataset collected in an off-road environment. It contains 13,556 annotated LiDAR scans, which present challenges to class imbalance and environmental topography. \textit{SemanticPOSS} \cite{pan2020semanticPOSS} is a small-scale point cloud dataset with rich dynamic instances captured in Peking University. It consists of $6$ LiDAR sequences, where sequence $2$ is the validation set and the remaining data forms the training set. \textit{SemanticSTF} \cite{xiao2023semanticSTF} consists of 2,076 LiDAR scans from various adverse weather conditions, including ``snowy'', ``dense-foggy'', ``light-foggy'', and ``rainy'' scans. The dataset is split into three sets: 1,326 scans for training, 250 scans for validation, and 500 scans for testing. \textit{SynLiDAR} \cite{xiao2022synLiDAR}, \textit{Synth4D} \cite{saltori2020synth4D}, and \textit{DAPS-3D} \cite{klokov2023daps3D} are synthetic datasets captured from various simulators. \textit{SynLiDAR} \cite{xiao2022synLiDAR} contains $13$ LiDAR sequences with totally 198,396 samples. \textit{Synth4D} \cite{saltori2020synth4D} includes two subsets and we use Synth4D-nuScenes in this work. It comprises of 20,000 point clouds captured in different scenarios, including town, highway, rural area, and city. \textit{DAPS-3D} includes two subsets and we use DAPS-1, which is semi-synthetic with larger scale in this work. It contains $11$ sequences with about 23,000 LiDAR scans.

\noindent\textbf{Out-of-Distribution Robustness Evaluation.}
In this work, we conduct a comprehensive out-of-distribution (OoD) robustness evaluation experiment on the \textit{nuScenes-C} dataset from the Robo3D \cite{kong2023robo3D} benchmark. As shown in \cref{tab:robo3d_dataset}, there are a total of 8 OoD scenarios in the \textit{nuScenes-C} dataset, including ``fog'', ``wet ground'', ``snow'', ``motion blur'', ``beam missing'', ``crosstalk'', ``incomplete echo'', and ``cross sensor''. Each scenario is further split into three levels (``light'', ``moderate'', ``heavy'') based on its severity. We test each model on all three levels and report the average results.

\subsection{Training Configurations}
\label{subsec:training}

In this work, we implement the MinkUNet \cite{choy2019minkowski} network with the TorchSparse \cite{torchsparse} backend as our 3D backbone. The point clouds are partitioned under cylindrical voxels of size 0.10 meter. For the 3D network, point clouds are randomly rotated around the $z$-axis, flipped along $x$-axis and $y$-axis with a $50\%$ probability, and scaled with a factor between $0.95$ and $1.05$ during pretraining and downstream fine-tuning. For the 2D network, we choose ViT pretrained from DINOV2 \cite{oquab2023dinov2} with three variants: ViT-S, ViT-B, and ViT-L. The image data are resized to $224 \times 448$, and flipped horizontally with a $50\%$ probability during pretraining. For pretraining, we randomly choose $3$ camera images as inputs of the 2D network.
To enable view consistency alignment, we use the class names as the prompts when generating the semantic superpixels.
We train the network with eight GPUs for 50 epochs and the batch size is set to 4 for each GPU. For downstream fine-tuning, we use the same data split as \cite{seal} for all datasets. The loss function of segmentation is a combination of cross-entropy loss and Lov{\'a}sz-Softmax loss \cite{berman2018lovasz}. We train the segmentation network with four GPUs for 100 epochs and the batch size is set to 2 for each GPU. All the models are trained with the AdamW optimizer \cite{AdamW} and OneCycle scheduler \cite{OneCycle}. The learning rate is set as $0.01$ and $0.001$ for pretraining and fine-tuning, respectively.

\subsection{Evaluation Configurations}
\label{subsec:evaluation}

Following conventions, we report Intersection-over-Union (IoU) for each category $i$ and mean IoU (mIoU) across all categories. IoU can be formulated as follows:
\begin{equation}
    \label{eq:iou}
    \text{IoU}_{i} = \frac{\text{TP}_{i}}{\text{TP}_{i} + \text{FP}_{i} + \text{FN}_{i}}~,
\end{equation}
where $\text{TP}_{i}, \text{FP}_{i}, \text{FN}_{i}$ are true positives, false positives, and false negatives for category $i$, respectively. For robust protocol, we utilize the Corruption Error (CE) and Resilience Rate (RR) metrics, following Robo3D \cite{kong2023robo3D}, which are defined as follows:
\begin{equation}
    \label{eq:robo}
    \text{CE}_{i} = \frac{\sum_{j=1}^{3}(1-\text{IoU}_{i}^{j})}{\sum_{j=1}^{3}(1-\text{IoU}_{i_{\text{base}}}^{j})}~,
    \text{RR}_{i} = \frac{\sum_{j=1}^{3}(1-\text{IoU}_{i}^{j})}{3 \times \text{IoU}_{\text{clean}}}~,
\end{equation}
where $\text{IoU}_{i}^{j}$ is the mIoU calculated at the $i$-th scene for the $j$-th level; $\text{IoU}_{i_{\text{base}}}^{j}$ and $\text{IoU}_{\text{clean}}$ are scores of the baseline model and scores on the ``clean'' validation set. For a fair comparison with priors, all models are tested without test time augmentation or model ensemble for both linear probing and downstream tasks.

\begin{table}[t]
\centering
\caption{The \textbf{per-class IoU scores} of state-of-the-art pretraining methods pretrained and linear-probed on the \textit{nuScenes} \cite{Panoptic-nuScenes} dataset. All IoU scores are given in percentage (\%). The \textbf{best} IoU scores in each configuration are shaded with colors.}
\vspace{-0.2cm}
\label{tab:linear_probing}
\resizebox{\textwidth}{!}{
\begin{tabular}{r|c|cccccccccccccccc}
\toprule
\textbf{Method} & \rotatebox{90}{\textbf{mIoU}} & \rotatebox{90}{\textcolor{nu_barrier}{$\blacksquare$}~barrier} & \rotatebox{90}{\textcolor{nu_bicycle}{$\blacksquare$}~bicycle} & \rotatebox{90}{\textcolor{nu_bus}{$\blacksquare$}~bus}  & \rotatebox{90}{\textcolor{nu_car}{$\blacksquare$}~car}  & \rotatebox{90}{\textcolor{nu_cons}{$\blacksquare$}~construction vehicle} & \rotatebox{90}{\textcolor{nu_motor}{$\blacksquare$}~motorcycle} & \rotatebox{90}{\textcolor{nu_ped}{$\blacksquare$}~pedestrian} & \rotatebox{90}{\textcolor{nu_cone}{$\blacksquare$}~traffic cone} & \rotatebox{90}{\textcolor{nu_trailer}{$\blacksquare$}~trailer} & \rotatebox{90}{\textcolor{nu_truck}{$\blacksquare$}~truck} & \rotatebox{90}{\textcolor{nu_driv}{$\blacksquare$}~driveable surface} & \rotatebox{90}{\textcolor{nu_flat}{$\blacksquare$}~other flat} & \rotatebox{90}{\textcolor{nu_sidewalk}{$\blacksquare$}~sidewalk} & \rotatebox{90}{\textcolor{nu_terrain}{$\blacksquare$}~terrain} & \rotatebox{90}{\textcolor{nu_manmade}{$\blacksquare$}~manmade} & \rotatebox{90}{\textcolor{nu_veg}{$\blacksquare$}~vegetation}
\\\midrule\midrule
\textcolor{gray}{Random} & \textcolor{gray}{$8.1$} & \textcolor{gray}{$0.5$} & \textcolor{gray}{$0.0$} & \textcolor{gray}{$0.0$} & \textcolor{gray}{$3.9$} & \textcolor{gray}{$0.0$} & \textcolor{gray}{$0.0$} & \textcolor{gray}{$0.0$} & \textcolor{gray}{$6.4$} & \textcolor{gray}{$0.0$} & \textcolor{gray}{$3.9$} & \textcolor{gray}{$59.6$} & \textcolor{gray}{$0.0$} & \textcolor{gray}{$0.1$} & \textcolor{gray}{$16.2$} & \textcolor{gray}{$30.6$} & \textcolor{gray}{$12.0$}
\\\midrule
\rowcolor{sf_gray!18}\multicolumn{18}{l}{\textcolor{sf_gray}{$\bullet$~\textbf{Distill: None}}}
\\
PointContrast \cite{pointcontrast} & $21.9$ & - & - & - & - & - & - & - & - & - & - & - & - & - & - & - & -
\\
DepthContrast \cite{depthcontrast} & \cellcolor{sf_gray!18}$22.1$ & - & - & - & - & - & - & - & - & - & - & - & - & - & - & - & -
\\
ALSO \cite{boulch2023also} & - & - & - & - & - & - & - & - & - & - & - & - & - & - & - & - & -
\\
BEVContrast \cite{sautier2023bevcontrast} & - & - & - & - & - & - & - & - & - & - & - & - & - & - & - & - & -
\\\midrule
\rowcolor{sf_blue!13}\multicolumn{18}{l}{\textcolor{sf_blue}{$\bullet$~\textbf{Distill: ResNet-50}}}
\\
PPKT \cite{ppkt} & $35.9$ & - & - & - & - & - & - & - & - & - & - & - & - & - & - & - & -
\\
SLidR \cite{slidr} & $39.2$ & $44.2$ & $0.0$  & \cellcolor{sf_blue!13}$30.8$ & $60.2$ & $15.1$ & $22.4$  & $47.2$ & $27.7$ & $16.3$ & $34.3$ & $80.6$ & $21.8$  & $35.2$ &  $48.1$ & $71.0$ & $71.9$
\\
ST-SLidR \cite{st-slidr} & $40.5$ & - & - & - & - & - & - & - & - & - & - & - & - & - & - & - & -
\\
TriCC \cite{pang2023tricc} & $38.0$ & - & - & - & - & - & - & - & - & - & - & - & - & - & - & - & -
\\
Seal \cite{seal} & \cellcolor{sf_blue!13}$45.0$ & \cellcolor{sf_blue!13}$54.7$ & \cellcolor{sf_blue!13}$5.9$ & $30.6$ & \cellcolor{sf_blue!13}$61.7$ & \cellcolor{sf_blue!13}$18.9$ & \cellcolor{sf_blue!13}$28.8$ & \cellcolor{sf_blue!13}$48.1$ & \cellcolor{sf_blue!13}$31.0$ & \cellcolor{sf_blue!13}$22.1$ & \cellcolor{sf_blue!13}$39.5$ & \cellcolor{sf_blue!13}$83.8$ & \cellcolor{sf_blue!13}$35.4$ & \cellcolor{sf_blue!13}$46.7$ & \cellcolor{sf_blue!13}$56.9$ & \cellcolor{sf_blue!13}$74.7$ & \cellcolor{sf_blue!13}$74.7$ 
\\
HVDistill \cite{zhang2024hvdistill} & $39.5$ & - & - & - & - & - & - & - & - & - & - & - & - & - & - & - & -
\\\midrule
\rowcolor{sf_red!10}\multicolumn{18}{l}{\textcolor{sf_red}{$\bullet$~\textbf{Distill: ViT-S}}}
\\
PPKT \cite{ppkt} & $38.6$ & $43.8$ & $0.0$ & $31.2$ & $53.1$ & $15.2$ & $0.0$ & $42.2$ & $16.5$ & $18.3$ & $33.7$ & $79.1$ & $37.2$ & $45.2$ & $52.7$ & $75.6$ & $74.3$
\\
SLidR \cite{slidr} & $44.7$ & $45.0$ & $8.2$ & $34.8$ & $58.6$ & \cellcolor{sf_red!10}$23.4$ & \cellcolor{sf_red!10}$40.2$ & $43.8$ & $19.0$ & $22.9$ & $40.9$ & $82.7$ & $38.3$ & \cellcolor{sf_red!10}$47.6$ & $53.9$ & \cellcolor{sf_red!10}$77.8$ & $77.9$
\\
Seal \cite{seal} & $45.2$ & $48.9$ & \cellcolor{sf_red!10}$8.4$ & $30.7$ & \cellcolor{sf_red!10}$68.1$ & $17.5$ & $37.7$ & $57.7$ & $17.9$ & $20.9$ & $40.4$ & $83.8$ & $36.6$ & $44.2$ & $54.5$ & $76.2$ & \cellcolor{sf_red!10}$79.3$
\\
\textbf{SuperFlow} & \cellcolor{sf_red!10}$46.4$ & \cellcolor{sf_red!10}$49.8$ & $6.8$ & \cellcolor{sf_red!10}$45.9$ & $63.4$ & $18.5$ & $31.0$ & \cellcolor{sf_red!10}$60.3$ & \cellcolor{sf_red!10}$28.1$ & \cellcolor{sf_red!10}$25.4$ & \cellcolor{sf_red!10}$47.4$ & \cellcolor{sf_red!10}$86.2$ & \cellcolor{sf_red!10}$38.4$ & $47.4$ & \cellcolor{sf_red!10}$56.7$ & $74.9$ & $77.8$
\\\midrule
\rowcolor{sf_yellow!12}\multicolumn{18}{l}{\textcolor{sf_yellow}{$\bullet$~\textbf{Distill: ViT-B}}}
\\
PPKT \cite{ppkt} & $40.0$ & $29.6$ & $0.0$ & $30.7$ & $55.8$ & $6.3$ & $22.4$ & $56.7$ & $18.1$ & \cellcolor{sf_yellow!12}$24.3$ & $42.7$ & $82.3$ & $33.2$ & $45.1$ & $53.4$ & $71.3$ & $75.7$
\\
SLidR \cite{slidr} & $45.4$ & $46.7$ & $7.8$ & $46.5$ & $58.7$ & \cellcolor{sf_yellow!12}$23.9$ & $34.0$ & $47.8$ & $17.1$ & $23.7$ & $41.7$ & $83.4$ & \cellcolor{sf_yellow!12}$39.4$ & $47.0$ & $54.6$ & $76.6$ & $77.8$
\\
Seal \cite{seal} & $46.6$ & \cellcolor{sf_yellow!12}$49.3$ & $8.2$ & $35.1$ & \cellcolor{sf_yellow!12}$70.8$ & $22.1$ & \cellcolor{sf_yellow!12}$41.7$ & $57.4$ & $15.2$ & $21.6$ & $42.6$ & $84.5$ & $38.1$ & $46.8$ & $55.4$ & \cellcolor{sf_yellow!12}$77.2$ & \cellcolor{sf_yellow!12}$79.5$
\\
\textbf{SuperFlow} & \cellcolor{sf_yellow!12}$47.7$ & $45.8$ & \cellcolor{sf_yellow!12}$12.4$ & \cellcolor{sf_yellow!12}$52.6$ & $67.9$ & $17.2$ & $40.8$ & \cellcolor{sf_yellow!12}$59.5$ & \cellcolor{sf_yellow!12}$25.4$ & $21.0$ & \cellcolor{sf_yellow!12}$47.6$ & \cellcolor{sf_yellow!12}$85.8$ & $37.2$ & \cellcolor{sf_yellow!12}$48.4$ & \cellcolor{sf_yellow!12}$56.6$ & $76.2$ & $78.2$
\\\midrule
\rowcolor{sf_green!12}\multicolumn{18}{l}{\textcolor{sf_green}{$\bullet$~\textbf{Distill: ViT-L}}}
\\
PPKT \cite{ppkt} & $41.6$ & $30.5$ & $0.0$ & $32.0$ & $57.3$ & $8.7$ & $24.0$ & $58.1$ & $19.5$ & \cellcolor{sf_green!12}$24.9$ & $44.1$ & $83.1$ & $34.5$ & $45.9$ & $55.4$ & $72.5$ & $76.4$
\\
SLidR \cite{slidr} & $45.7$ & $46.9$ & $6.9$ & $44.9$ & $60.8$ & $22.7$ & $40.6$ & $44.7$ & $17.4$ & $23.0$ & $40.4$ & $83.6$ & \cellcolor{sf_green!12}$39.9$ & \cellcolor{sf_green!12}$47.8$ & $55.2$ & $78.1$ & $78.3$
\\
Seal \cite{seal} & $46.8$ & $53.1$ & $6.9$ & $35.0$ & $65.0$ & $22.0$ & $46.1$ & \cellcolor{sf_green!12}$59.2$ & $16.2$ & $23.0$ & $41.8$ & $84.7$ & $35.8$ & $46.6$ & $55.5$ & \cellcolor{sf_green!12}$78.4$ & \cellcolor{sf_green!12}$79.8$
\\
\textbf{SuperFlow} & \cellcolor{sf_green!12}$48.0$ & \cellcolor{sf_green!12}$54.1$ & \cellcolor{sf_green!12}$14.9$ & \cellcolor{sf_green!12}$47.6$ & \cellcolor{sf_green!12}$65.9$ & \cellcolor{sf_green!12}$23.4$ & \cellcolor{sf_green!12}$46.5$ & $56.9$ & \cellcolor{sf_green!12}$27.5$ & $20.7$ & \cellcolor{sf_green!12}$44.4$ & \cellcolor{sf_green!12}$84.8$ & $39.2$ & $47.4$ & \cellcolor{sf_green!12}$58.0$ & $76.0$ & $79.2$
\\\bottomrule
\end{tabular}}
\vspace{-0.2cm}
\end{table}

\begin{table}[t]
\centering
\caption{The \textbf{per-class IoU scores} of state-of-the-art pretraining methods pretrained and fine-tuned on \textit{nuScenes} \cite{Panoptic-nuScenes} with $1\%$ annotations. All IoU scores are given in percentage (\%). The \textbf{best} IoU scores in each configuration are shaded with colors.}
\vspace{-0.2cm}
\label{tab:1pct}
\resizebox{\textwidth}{!}{
\begin{tabular}{r|c|cccccccccccccccc}
\toprule
\textbf{Method} & \rotatebox{90}{\textbf{mIoU}} & \rotatebox{90}{\textcolor{nu_barrier}{$\blacksquare$}~barrier} & \rotatebox{90}{\textcolor{nu_bicycle}{$\blacksquare$}~bicycle} & \rotatebox{90}{\textcolor{nu_bus}{$\blacksquare$}~bus}  & \rotatebox{90}{\textcolor{nu_car}{$\blacksquare$}~car}  & \rotatebox{90}{\textcolor{nu_cons}{$\blacksquare$}~construction vehicle} & \rotatebox{90}{\textcolor{nu_motor}{$\blacksquare$}~motorcycle} & \rotatebox{90}{\textcolor{nu_ped}{$\blacksquare$}~pedestrian} & \rotatebox{90}{\textcolor{nu_cone}{$\blacksquare$}~traffic cone} & \rotatebox{90}{\textcolor{nu_trailer}{$\blacksquare$}~trailer} & \rotatebox{90}{\textcolor{nu_truck}{$\blacksquare$}~truck} & \rotatebox{90}{\textcolor{nu_driv}{$\blacksquare$}~driveable surface} & \rotatebox{90}{\textcolor{nu_flat}{$\blacksquare$}~other flat} & \rotatebox{90}{\textcolor{nu_sidewalk}{$\blacksquare$}~sidewalk} & \rotatebox{90}{\textcolor{nu_terrain}{$\blacksquare$}~terrain} & \rotatebox{90}{\textcolor{nu_manmade}{$\blacksquare$}~manmade} & \rotatebox{90}{\textcolor{nu_veg}{$\blacksquare$}~vegetation}
\\\midrule\midrule
\textcolor{gray}{Random} & \textcolor{gray}{$30.3$} & \textcolor{gray}{$0.0$} & \textcolor{gray}{$0.0$} & \textcolor{gray}{$8.1$} & \textcolor{gray}{$65.0$} & \textcolor{gray}{$0.1$} & \textcolor{gray}{$6.6$} & \textcolor{gray}{$21.0$} & \textcolor{gray}{$9.0$} & \textcolor{gray}{$9.3$} & \textcolor{gray}{$25.8$} & \textcolor{gray}{$89.5$} & \textcolor{gray}{$14.8$} & \textcolor{gray}{$41.7$} & \textcolor{gray}{$48.7$} & \textcolor{gray}{$72.4$} & \textcolor{gray}{$73.3$}
\\\midrule
\rowcolor{sf_gray!18}\multicolumn{18}{l}{\textcolor{sf_gray}{$\bullet$~\textbf{Distill: None}}}
\\
PointContrast \cite{pointcontrast} & $32.5$ & $0.0$ & $1.0$ & $5.6$ & $67.4$ & $0.0$ & $3.3$ & $31.6$ & $5.6$ & $12.1$ & $30.8$ & \cellcolor{sf_gray!18}$91.7$ & $21.9$ & \cellcolor{sf_gray!18}$48.4$ & $50.8$ & $75.0$ & $74.6$
\\
DepthContrast \cite{depthcontrast} & $31.7$ & $0.0$ & $0.6$ & $6.5$ & $64.7$ & $0.2$ & \cellcolor{sf_gray!18}$5.1$ & $29.0$ & \cellcolor{sf_gray!18}$9.5$ & $12.1$ & $29.9$ & $90.3$ & $17.8$ & $44.4$ & $49.5$ & $73.5$ & $74.0$
\\
ALSO \cite{boulch2023also} & $37.7$ & - & - & - & - & - & - & - & - & - & - & - & - & - & - & - & -
\\
BEVContrast \cite{sautier2023bevcontrast} & \cellcolor{sf_gray!18}$37.9$ & $0.0$ & \cellcolor{sf_gray!18}$1.3$ & \cellcolor{sf_gray!18}$32.6$ & \cellcolor{sf_gray!18}$74.3$ & \cellcolor{sf_gray!18}$1.1$ & $0.9$ & \cellcolor{sf_gray!18}$41.3$ & $8.1$ & \cellcolor{sf_gray!18}$24.1$ & \cellcolor{sf_gray!18}$40.9$ & $89.8$ & \cellcolor{sf_gray!18}$36.2$ & $44.0$ & \cellcolor{sf_gray!18}$52.1$ & \cellcolor{sf_gray!18}$79.9$ & \cellcolor{sf_gray!18}$79.7$
\\\midrule
\rowcolor{sf_blue!13}\multicolumn{18}{l}{\textcolor{sf_blue}{$\bullet$~\textbf{Distill: ResNet-50}}}
\\
PPKT \cite{ppkt} & $37.8$ & $0.0$ & $2.2$ & $20.7$ & $75.4$ & $1.2$ & $13.2$ & $45.6$ & $8.5$ & $17.5$ & $38.4$ & $92.5$ & $19.2$ & $52.3$ & $56.8$ & $80.1$ & $80.9$
\\
SLidR \cite{slidr} & $38.8$ & $0.0$ & $1.8$ & $15.4$ & $73.1$ & $1.9$ 
& $19.9$ & $47.2$ & $17.1$ & $14.5$ & $34.5$ & $92.0$ & $27.1$ & $53.6$ & \cellcolor{sf_blue!13}$61.0$ & $79.8$ & $82.3$
\\
ST-SLidR \cite{st-slidr} & $40.8$ & - & - & - & - & - & - & - & - & - & - & - & - & - & - & - & -
\\
TriCC \cite{pang2023tricc} & $41.2$ & - & - & - & - & - & - & - & - & - & - & - & - & - & - & - & -
\\
Seal \cite{seal} & \cellcolor{sf_blue!13}$45.8$ & $0.0$ & \cellcolor{sf_blue!13}$9.4$ & \cellcolor{sf_blue!13}$32.6$ & \cellcolor{sf_blue!13}$77.5$ & \cellcolor{sf_blue!13}$10.4$ & \cellcolor{sf_blue!13}$28.0$ & \cellcolor{sf_blue!13}$53.0$ & \cellcolor{sf_blue!13}$25.0$ & \cellcolor{sf_blue!13}$30.9$ & \cellcolor{sf_blue!13}$49.7$ & \cellcolor{sf_blue!13}$94.0$ & \cellcolor{sf_blue!13}$33.7$ & \cellcolor{sf_blue!13}$60.1$ & $59.6$ & \cellcolor{sf_blue!13}$83.9$ & \cellcolor{sf_blue!13}$83.4$
\\
HVDistill \cite{zhang2024hvdistill} & $42.7$ & - & - & - & - & - & - & - & - & - & - & - & - & - & - & - & -
\\\midrule
\rowcolor{sf_red!10}\multicolumn{18}{l}{\textcolor{sf_red}{$\bullet$~\textbf{Distill: ViT-S}}}
\\
PPKT \cite{ppkt} & $40.6$ & $0.0$ & $0.0$ & $25.2$ & $73.5$ & $9.1$ & $6.9$ & $51.4$ & $8.6$ & $11.3$ & $31.1$ & $93.2$ & \cellcolor{sf_red!10}$41.7$ & $58.3$ & $64.0$ & $82.0$ & $82.6$
\\
SLidR \cite{slidr} & $41.2$ & $0.0$ & $0.0$ & \cellcolor{sf_red!10}$26.6$ & $72.0$ & $12.4$ & $15.8$ & $51.4$ & $22.9$ & $11.7$ & $35.3$ & $92.9$ & $36.3$ & $58.7$ & $63.6$ & $81.2$ & $82.3$
\\
Seal \cite{seal} & $44.3$ & $20.0$ & $0.0$ & $19.4$ & $74.7$ & $10.6$ & \cellcolor{sf_red!10}$45.7$ & \cellcolor{sf_red!10}$60.3$ & \cellcolor{sf_red!10}$29.2$ & $17.4$ & $38.1$ & $93.2$ & $26.0$ & $58.8$ & \cellcolor{sf_red!10}$64.5$ & $81.9$ & $81.9$
\\
\textbf{SuperFlow} & \cellcolor{sf_red!10}$47.8$ & \cellcolor{sf_red!10}$38.2$ & \cellcolor{sf_red!10}$1.8$ & $25.8$ & \cellcolor{sf_red!10}$79.0$ & \cellcolor{sf_red!10}$15.3$ & $43.6$ & \cellcolor{sf_red!10}$60.3$ & $0.0$ & \cellcolor{sf_red!10}$28.4$ & \cellcolor{sf_red!10}$55.4$ & \cellcolor{sf_red!10}$93.7$ & $28.8$ & \cellcolor{sf_red!10}$59.1$ & $59.9$ & \cellcolor{sf_red!10}$83.5$ & \cellcolor{sf_red!10}$83.1$
\\\midrule
\rowcolor{sf_yellow!12}\multicolumn{18}{l}{\textcolor{sf_yellow}{$\bullet$~\textbf{Distill: ViT-B}}}
\\
PPKT \cite{ppkt} & $40.9$ & $0.0$ & $0.0$ & $24.5$ & $73.5$ & \cellcolor{sf_yellow!12}$12.2$ & $7.0$ & $51.0$ & $13.5$ & $15.4$ & $36.3$ & $93.1$ & $40.4$ & $59.2$ & $63.5$ & $81.7$ & $82.2$
\\
SLidR \cite{slidr} & $41.6$ & $0.0$ & $0.0$ & $26.7$ & $73.4$ & $10.3$ & $16.9$ & $51.3$ & \cellcolor{sf_yellow!12}$23.3$ & $12.7$ & $38.1$ & $93.0$ & $37.7$ & $58.8$ & $63.4$ & $81.6$ & $82.7$
\\
Seal \cite{seal} & $46.0$ & \cellcolor{sf_yellow!12}$43.0$ & $0.0$ & $26.7$ & \cellcolor{sf_yellow!12}$81.3$ & $9.9$ & $41.3$ & $56.2$ & $0.0$ & $21.7$ & $51.6$ & $93.6$ & \cellcolor{sf_yellow!12}$42.3$ & \cellcolor{sf_yellow!12}$62.8$ & \cellcolor{sf_yellow!12}$64.7$ & $82.6$ & $82.7$
\\
\textbf{SuperFlow} & \cellcolor{sf_yellow!12}$48.1$ & $39.1$ & \cellcolor{sf_yellow!12}$0.9$ & \cellcolor{sf_yellow!12}$30.0$ & $80.7$ & $10.3$ & \cellcolor{sf_yellow!12}$47.1$ & \cellcolor{sf_yellow!12}$59.5$ & $5.1$ & \cellcolor{sf_yellow!12}$27.6$ & \cellcolor{sf_yellow!12}$55.4$ & \cellcolor{sf_yellow!12}$93.7$ & $29.1$ & $61.1$ & $63.5$ & \cellcolor{sf_yellow!12}$82.7$ & \cellcolor{sf_yellow!12}$83.6$
\\\midrule
\rowcolor{sf_green!12}\multicolumn{18}{l}{\textcolor{sf_green}{$\bullet$~\textbf{Distill: ViT-L}}}
\\
PPKT \cite{ppkt} & $42.1$ & $0.0$ & $0.0$ & \cellcolor{sf_green!12}$24.4$ & $78.8$ & $15.1$ & $9.2$ & $54.2$ & $14.3$ & $12.9$ & $39.1$ & $92.9$ & $37.8$ & $59.8$ & $64.9$ & $82.3$ & $83.6$
\\
SLidR \cite{slidr} & $42.8$ & $0.0$ & $0.0$ & $23.9$ & $78.8$ & $15.2$ & $20.9$ & $55.0$ & \cellcolor{sf_green!12}$28.0$ & $17.4$ & $41.4$ & $92.2$ & $41.2$ & $58.0$ & $64.0$ & $81.8$ & $82.7$
\\
Seal \cite{seal} & $46.3$ & $41.8$ & $0.0$ & $23.8$ & \cellcolor{sf_green!12}$81.4$ & \cellcolor{sf_green!12}$17.7$ & $46.3$ & $58.6$ & $0.0$ & $23.4$ & $54.7$ & $93.8$ & \cellcolor{sf_green!12}$41.4$ & \cellcolor{sf_green!12}$62.5$ & \cellcolor{sf_green!12}$65.0$ & \cellcolor{sf_green!12}$83.8$ & \cellcolor{sf_green!12}$83.8$
\\
\textbf{SuperFlow} & \cellcolor{sf_green!12}$50.0$ & \cellcolor{sf_green!12}$44.5$ & \cellcolor{sf_green!12}$0.9$ & $22.4$ & $80.8$ & $17.1$ & \cellcolor{sf_green!12}$50.2$ & \cellcolor{sf_green!12}$60.9$ & $21.0$ & \cellcolor{sf_green!12}$25.1$ & \cellcolor{sf_green!12}$55.1$ & \cellcolor{sf_green!12}$93.9$ & $35.8$ & $61.5$ & $62.6$ & $83.7$ & $83.7$
\\\bottomrule
\end{tabular}}
\vspace{-0.2cm}
\end{table}

\section{Additional Quantitative Result}
\label{sec:supp_quantitative}

In this section, we supplement the complete results (\ie, the class-wise LiDAR semantic segmentation results) to better support the findings and conclusions drawn in the main body of this paper.

\subsection{Class-Wise Linear Probing Results}
\label{subsec:linear_probe}

We present the class-wise IoU scores for the linear probing experiments in \cref{tab:linear_probing}. We also implement PPKT \cite{ppkt}, SLidR \cite{slidr}, and Seal \cite{seal} with the distillation of ViT-S, ViT-B, and ViT-L. The results show that SuperFlow outperforms state-of-the-art pretraining methods significantly for most semantic classes.
Some notably improved classes are: ``barrier'', ``bus'', ``traffic cone'', and ``terrain''. Additionally, we observe a consistent trend of performance improvements using larger models for the cross-sensor distillation.

\subsection{Class-Wise Fine-Tuning Results}
\label{subsec:fine_tune}

We present the class-wise IoU scores for the 1\% fine-tuning experiments in \cref{tab:1pct}. We observe that a holistic improvement brought by SuperFlow compared to state-of-the-art pretraining methods.

\section{Additional Qualitative Result}
\label{sec:supp_qualitative}

In this section, we provide additional qualitative examples to help visually compare different approaches presented in the main body of this paper.

\subsection{LiDAR Segmentation Results}
\label{subsec:lidarseg}

\begin{figure}[t]
    \begin{center}
    \includegraphics[width=\textwidth]{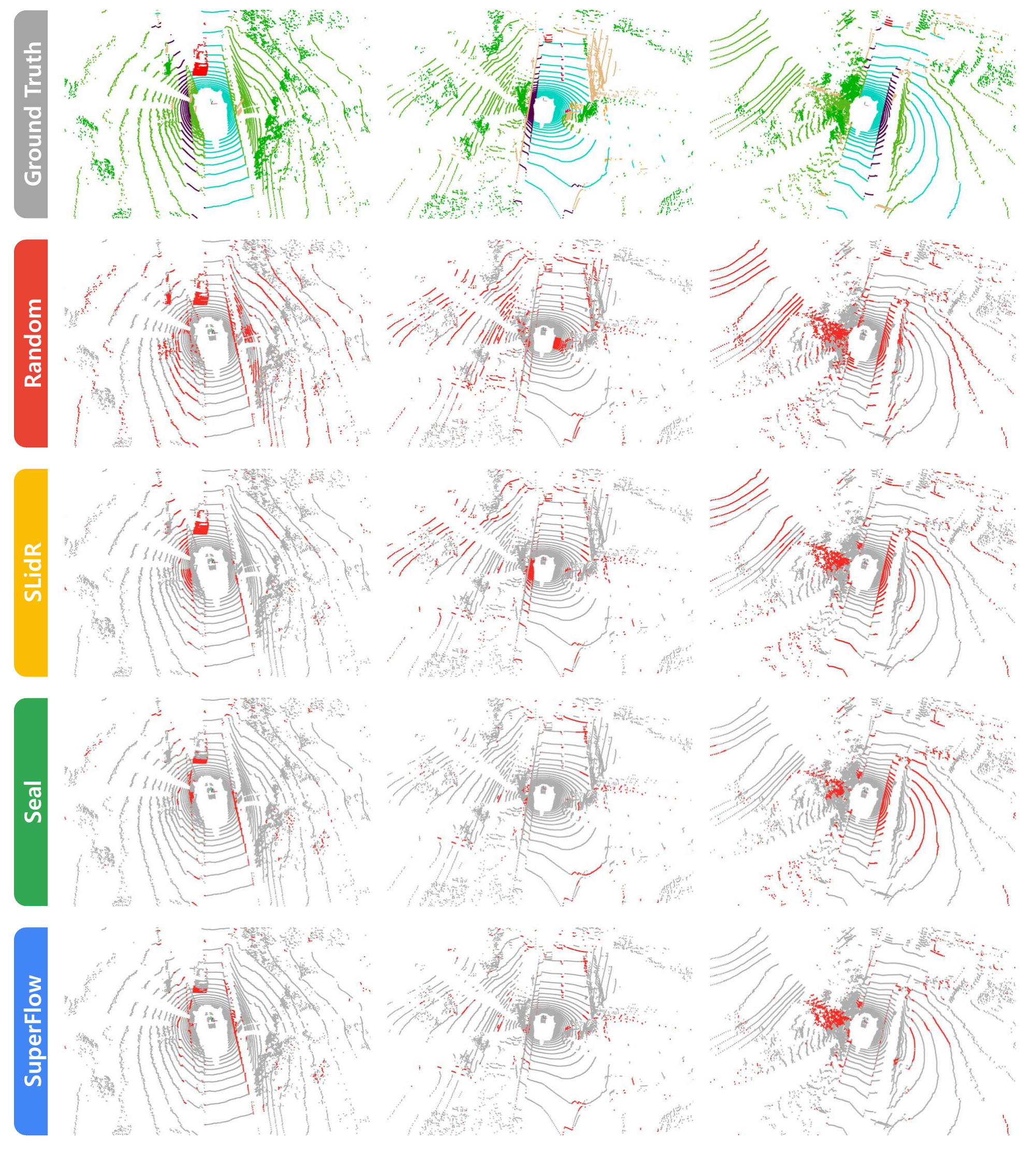}
    \end{center}
    \vspace{-0.35cm}
    \caption{\textbf{Qualitative assessments} of state-of-the-art pretraining methods pretrained on \textit{nuScenes} \cite{Panoptic-nuScenes} and fine-tuned on \textit{nuScenes} \cite{Panoptic-nuScenes} with $1\%$ annotations. The error maps show the \textcolor{gray}{correct} and \textcolor{sf_red}{incorrect} predictions in \textcolor{gray}{gray} and \textcolor{sf_red}{red}, respectively. Best viewed in colors and zoomed-in for details.}
    \label{fig:supp_qua_nusc}
\end{figure}

\begin{figure}[t]
    \begin{center}
    \includegraphics[width=\textwidth]{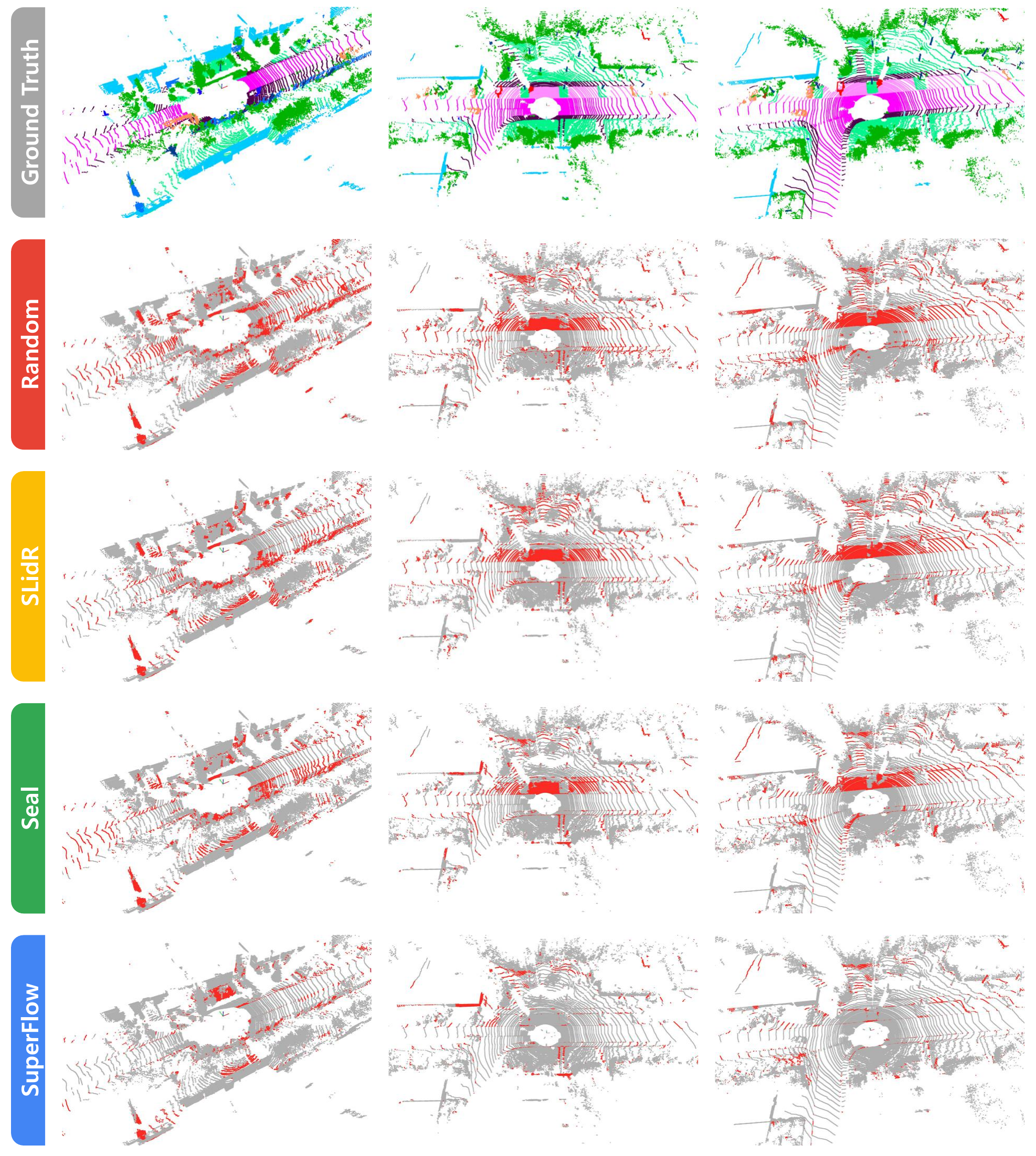}
    \end{center}
    \vspace{-0.35cm}
    \caption{\textbf{Qualitative assessments} of state-of-the-art pretraining methods pretrained on \textit{nuScenes} \cite{Panoptic-nuScenes} and fine-tuned on \textit{SemanticKITTI} \cite{SemanticKITTI} with $1\%$ annotations. The error maps show the \textcolor{gray}{correct} and \textcolor{sf_red}{incorrect} predictions in \textcolor{gray}{gray} and \textcolor{sf_red}{red}, respectively. Best viewed in colors and zoomed-in for details.}
    \label{fig:supp_qua_semkitti}
\end{figure}

\begin{figure}[t]
    \begin{center}
    \includegraphics[width=\textwidth]{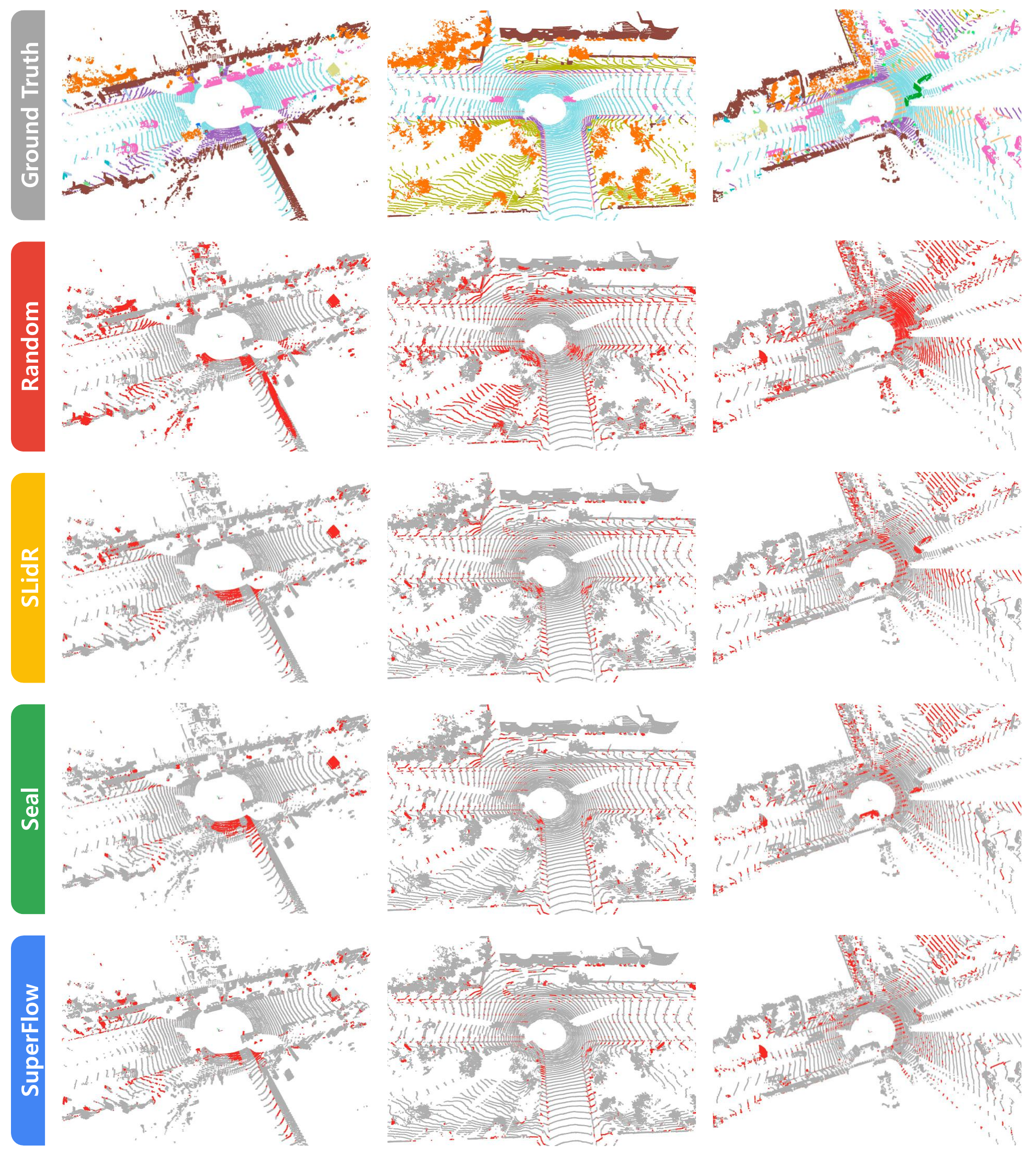}
    \end{center}
    \vspace{-0.35cm}
    \caption{\textbf{Qualitative assessments} of state-of-the-art pretraining methods pretrained on \textit{nuScenes} \cite{Panoptic-nuScenes} and fine-tuned on \textit{Waymo Open} \cite{sun2020waymoOpen} with $1\%$ annotations. The error maps show the \textcolor{gray}{correct} and \textcolor{sf_red}{incorrect} predictions in \textcolor{gray}{gray} and \textcolor{sf_red}{red}, respectively. Best viewed in colors and zoomed-in for details.}
    \label{fig:supp_qua_waymo}
\end{figure}

We provide additional qualitative assessments in \cref{fig:supp_qua_nusc}, \cref{fig:supp_qua_semkitti}, and \cref{fig:supp_qua_waymo}. The results verify again the superiority of SuperFlow over prior pretraining methods.

\subsection{Cosine Similarity Results}
\label{subsec:cosine_similarity}

\begin{figure}[t]
    \centering
    \begin{subfigure}[b]{0.45\textwidth}
        \centering
        \includegraphics[width=\textwidth]{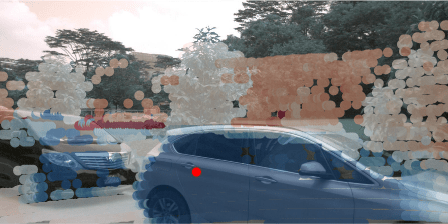}
        \caption{``car'' (3D)}
        \label{fig:headmap_3d_1}
    \end{subfigure}
    ~~
    \centering
    \begin{subfigure}[b]{0.45\textwidth}
        \centering
        \includegraphics[width=\textwidth]{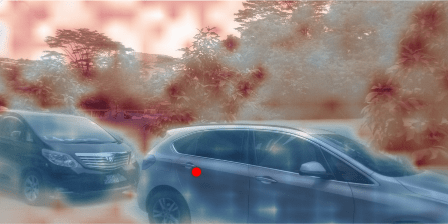}
        \caption{``car'' (2D)}
        \label{fig:headmap_2d_1}
    \end{subfigure}
    \centering
    \begin{subfigure}[b]{0.45\textwidth}
        \centering
        \includegraphics[width=\textwidth]{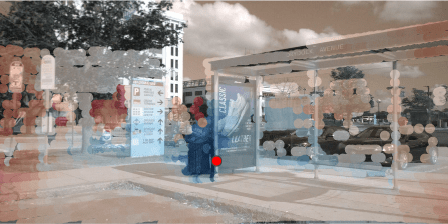}
        \caption{``flat-other'' (3D)}
        \label{fig:headmap_3d_2}
    \end{subfigure}
    ~~
    \centering
    \begin{subfigure}[b]{0.45\textwidth}
        \centering
        \includegraphics[width=\textwidth]{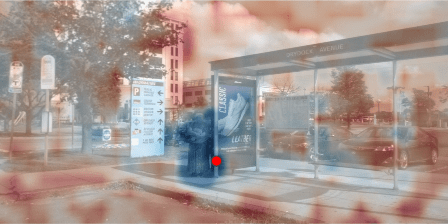}
        \caption{``flat-other'' (2D)}
        \label{fig:headmap_2d_2}
    \end{subfigure}
    \centering
    \begin{subfigure}[b]{0.45\textwidth}
        \centering
        \includegraphics[width=\textwidth]{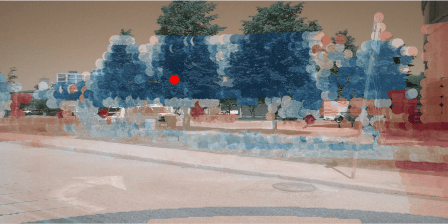}
        \caption{``terrain'' (3D)}
        \label{fig:headmap_3d_3}
    \end{subfigure}
    ~~
    \centering
    \begin{subfigure}[b]{0.45\textwidth}
        \centering
        \includegraphics[width=\textwidth]{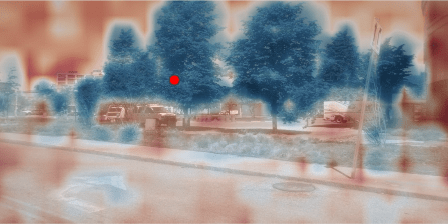}
        \caption{``terrain'' (2D)}
        \label{fig:headmap_2d_3}
    \end{subfigure}
    \centering
    \begin{subfigure}[b]{0.45\textwidth}
        \centering
        \includegraphics[width=\textwidth]{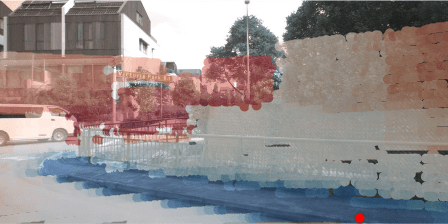}
        \caption{``sidewalk'' (3D)}
        \label{fig:headmap_3d_4}
    \end{subfigure}
    ~~
    \centering
    \begin{subfigure}[b]{0.45\textwidth}
        \centering
        \includegraphics[width=\textwidth]{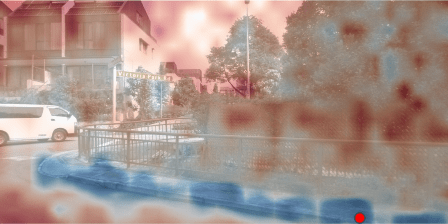}
        \caption{``sidewalk'' (2D)}
        \label{fig:headmap_2d_4}
    \end{subfigure}
    \centering
    \begin{subfigure}[b]{0.45\textwidth}
        \centering
        \includegraphics[width=\textwidth]{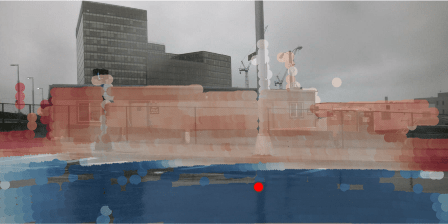}
        \caption{``driveable-surface'' (3D)}
        \label{fig:headmap_3d_5}
    \end{subfigure}
    ~~
    \centering
    \begin{subfigure}[b]{0.45\textwidth}
        \centering
        \includegraphics[width=\textwidth]{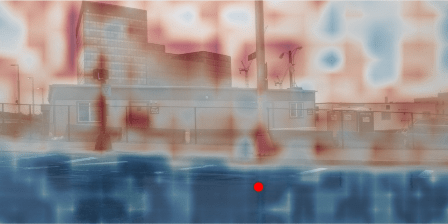}
        \caption{``driveable-surface'' (2D)}
        \label{fig:headmap_2d_5}
    \end{subfigure}
    
    \caption{\textbf{Cosine similarity} between the features of a query point (denoted as a \textcolor{red}{red dot}) and the features of other points projected in the image (the left column), and the features of an image with the same scene (the right column). The color goes from \textcolor{red}{red} to \textcolor{blue}{blue} denoting low and high similarity scores, respectively.}
    \label{fig:sim_supp1}
\end{figure}

\begin{figure}[t]
    \centering
    \begin{subfigure}[b]{0.45\textwidth}
        \centering
        \includegraphics[width=\textwidth]{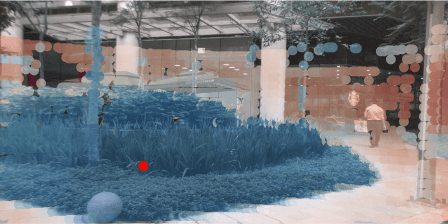}
        \caption{``vegetation'' (3D)}
        \label{fig:headmap_3d_6}
    \end{subfigure}
    ~~
    \centering
    \begin{subfigure}[b]{0.45\textwidth}
        \centering
        \includegraphics[width=\textwidth]{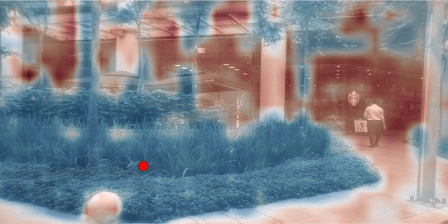}
        \caption{``vegetation'' (2D)}
        \label{fig:headmap_2d_6}
    \end{subfigure}
    \centering
    \begin{subfigure}[b]{0.45\textwidth}
        \centering
        \includegraphics[width=\textwidth]{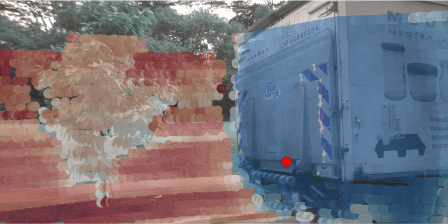}
        \caption{``construction-vehicle'' (3D)}
        \label{fig:headmap_3d_7}
    \end{subfigure}
    ~~
    \centering
    \begin{subfigure}[b]{0.45\textwidth}
        \centering
        \includegraphics[width=\textwidth]{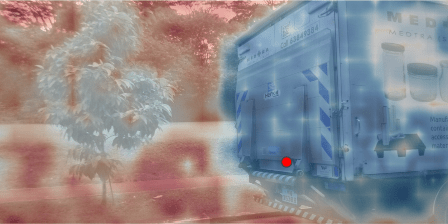}
        \caption{``construction-vehicle'' (2D)}
        \label{fig:headmap_2d_7}
    \end{subfigure}
    \centering
    \begin{subfigure}[b]{0.45\textwidth}
        \centering
        \includegraphics[width=\textwidth]{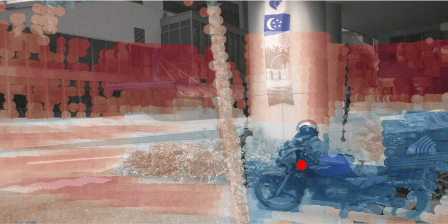}
        \caption{``motorcycle'' (3D)}
        \label{fig:headmap_3d_8}
    \end{subfigure}
    ~~
    \centering
    \begin{subfigure}[b]{0.45\textwidth}
        \centering
        \includegraphics[width=\textwidth]{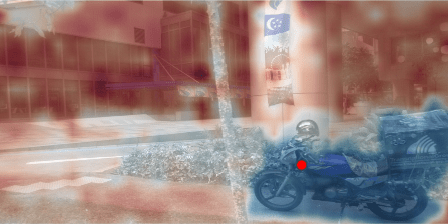}
        \caption{``motorcycle'' (2D)}
        \label{fig:headmap_2d_8}
    \end{subfigure}
    \centering
    \begin{subfigure}[b]{0.45\textwidth}
        \centering
        \includegraphics[width=\textwidth]{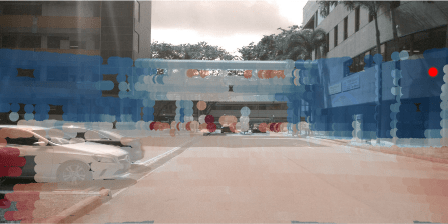}
        \caption{``manmade'' (3D)}
        \label{fig:headmap_3d_9}
    \end{subfigure}
    ~~
    \centering
    \begin{subfigure}[b]{0.45\textwidth}
        \centering
        \includegraphics[width=\textwidth]{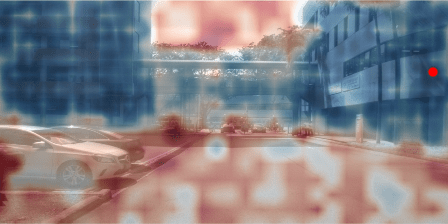}
        \caption{``manmade'' (2D)}
        \label{fig:headmap_2d_9}
    \end{subfigure}
    \centering
    \begin{subfigure}[b]{0.45\textwidth}
        \centering
        \includegraphics[width=\textwidth]{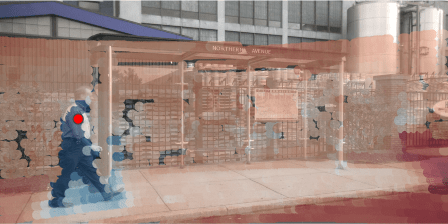}
        \caption{``pedestrian'' (3D)}
        \label{fig:headmap_3d_10}
    \end{subfigure}
    ~~
    \centering
    \begin{subfigure}[b]{0.45\textwidth}
        \centering
        \includegraphics[width=\textwidth]{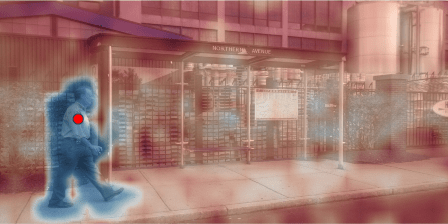}
        \caption{``pedestrian'' (2D)}
        \label{fig:headmap_2d_10}
    \end{subfigure}
    
    \caption{\textbf{Cosine similarity} between the features of a query point (denoted as a \textcolor{red}{red dot}) and the features of other points projected in the image (the left column), and the features of an image with the same scene (the right column). The color goes from \textcolor{red}{red} to \textcolor{blue}{blue} denoting low and high similarity scores, respectively.}
    \label{fig:sim_supp2}
\end{figure}

We provide additional cosine similarity maps in \cref{fig:sim_supp1} and \cref{fig:sim_supp2}. The results consistently verify the efficacy of SuperFlow in learning meaningful representations during flow-based spatiotemporal contrastive learning.

\section{Limitation and Discussion}
\label{sec:supp_limitation}

In this section, we elaborate on the limitations and potential negative societal impact of this work.

\subsection{Potential Limitations}
\label{subsec:potential_limitation}

Although SuperFlow holistically improves the image-to-LiDAR self-supervised learning efficacy, there are still rooms for further explorations.

\begin{wrapfigure}{r}{0.56\textwidth}
\vspace{-0.1cm}
    \centering
    \includegraphics[width=0.56\textwidth]{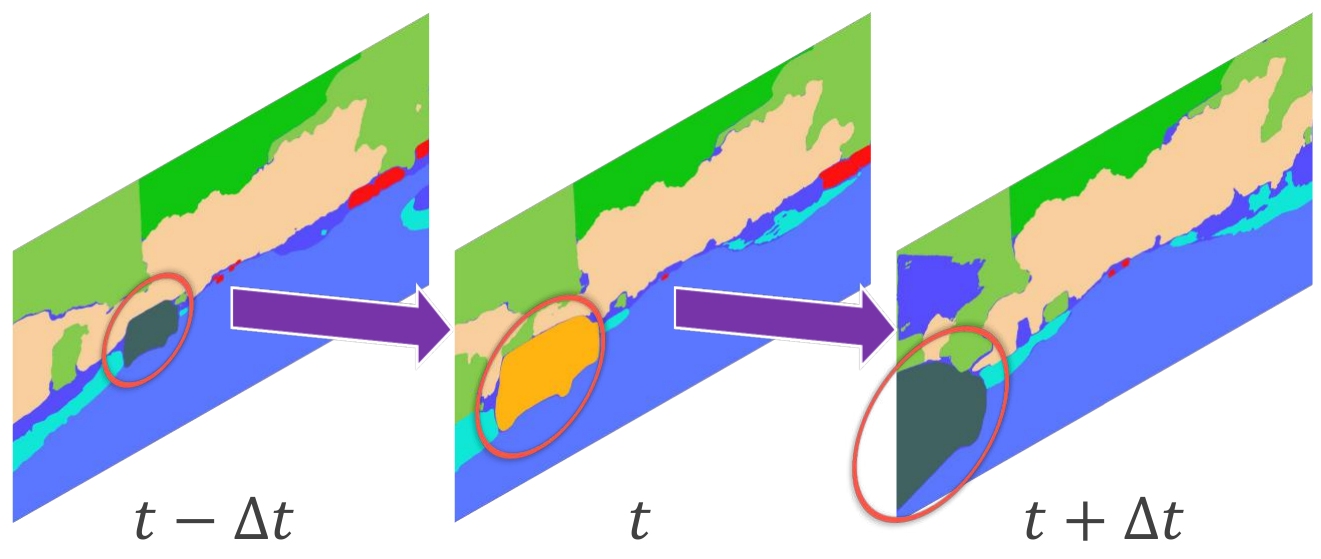}
    \vspace{-0.5cm}
    \caption{Possible temporal conflicts.}
    \label{fig:limination}
\vspace{-0.5cm}
\end{wrapfigure}

\noindent\textbf{Dynamic Objects.}
As shown in \cref{fig:limination}, dynamic objects across frames may share different superpixels due to variant scales in the images. The objects across frames will be regarded as negative samples, which will cause ``temporal conflict'' under temporal contrastive learning.

\noindent\textbf{Mis-Align Between LiDAR and Cameras.}
The calibration parameters between LiDAR and cameras are not perfect as they work at different frequencies. This will cause possible misalignment between superpoints and superpixels, especially when using dense point clouds to distill sparse point clouds. This also restricts the scalability to form much denser points from sweep points.

\subsection{Potential Societal Impact}
\label{subsec:potential_societal_impact}

LiDAR systems can capture detailed 3D images of environments, potentially including private spaces or sensitive information. If not properly managed, this could lead to privacy intrusions, as individuals might be identifiable from the data collected, especially when combined with other data sources.
Additionally, dependence on automated systems that use LiDAR semantic segmentation could lead to overreliance and trust in technology, potentially causing safety issues if the systems fail or make incorrect decisions. This is particularly critical in applications involving human safety.
\section{Public Resources Used}
\label{sec:supp_acknowledge}

In this section, we acknowledge the use of the following public resources, during the course of this work.

\subsection{Public Codebase Used}
\label{subsec:acknowledge_codebase}

We acknowledge the use of the following public codebase during this work:
\begin{itemize}
    \item MMCV\footnote{\url{https://github.com/open-mmlab/mmcv}.} \dotfill Apache License 2.0
    \item MMDetection\footnote{\url{https://github.com/open-mmlab/mmdetection}.} \dotfill Apache License 2.0
    \item MMDetection3D\footnote{\url{https://github.com/open-mmlab/mmdetection3d}.} \dotfill Apache License 2.0
    \item MMEngine\footnote{\url{https://github.com/open-mmlab/mmengine}.} \dotfill Apache License 2.0
    \item MMPreTrain\footnote{\url{https://github.com/open-mmlab/mmpretrain}.} \dotfill Apache License 2.0
    \item OpenPCSeg\footnote{\url{https://github.com/PJLab-ADG/OpenPCSeg}.} \dotfill Apache License 2.0
\end{itemize}

\subsection{Public Datasets Used}
\label{subsec:acknowledge_datasets}

We acknowledge the use of the following public datasets during this work:
\begin{itemize}
    \item nuScenes\footnote{\url{https://www.nuscenes.org/nuscenes}.} \dotfill CC BY-NC-SA 4.0
    \item nuScenes-devkit\footnote{\url{https://github.com/nutonomy/nuscenes-devkit}.} \dotfill Apache License 2.0
    \item SemanticKITTI\footnote{\url{http://semantic-kitti.org}.} \dotfill CC BY-NC-SA 4.0
    \item SemanticKITTI-API\footnote{\url{https://github.com/PRBonn/semantic-kitti-api}.} \dotfill MIT License
    \item WaymoOpenDataset\footnote{\url{https://waymo.com/open}.} \dotfill Waymo Dataset License
    \item Synth4D\footnote{\url{https://github.com/saltoricristiano/gipso-sfouda}.} \dotfill GPL-3.0 License
    \item ScribbleKITTI\footnote{\url{https://github.com/ouenal/scribblekitti}.} \dotfill Unknown
    \item RELLIS-3D\footnote{\url{https://github.com/unmannedlab/RELLIS-3D}.} \dotfill CC BY-NC-SA 3.0
    \item SemanticPOSS\footnote{\url{http://www.poss.pku.edu.cn/semanticposs.html}.} \dotfill CC BY-NC-SA 3.0
    \item SemanticSTF\footnote{\url{https://github.com/xiaoaoran/SemanticSTF}.} \dotfill CC BY-NC-SA 4.0
    \item SynthLiDAR\footnote{\url{https://github.com/xiaoaoran/SynLiDAR}.} \dotfill MIT License
    \item DAPS-3D\footnote{\url{https://github.com/subake/DAPS3D}.} \dotfill MIT License
    \item Robo3D\footnote{\url{https://github.com/ldkong1205/Robo3D}.} \dotfill CC BY-NC-SA 4.0
\end{itemize}

\subsection{Public Implementations Used}
\label{subsec:acknowledge_implements}

We acknowledge the use of the following implementations during this work:
\begin{itemize}
    \item SLidR\footnote{\url{https://github.com/valeoai/SLidR}.} \dotfill Apache License 2.0
    \item DINOv2\footnote{\url{https://github.com/facebookresearch/dinov2}.} \dotfill Apache License 2.0
    \item Segment-Any-Point-Cloud\footnote{\url{https://github.com/youquanl/Segment-Any-Point-Cloud}.} \dotfill CC BY-NC-SA 4.0
    \item OpenSeeD\footnote{\url{https://github.com/IDEA-Research/OpenSeeD}.} \dotfill Apache License 2.0
    \item torchsparse\footnote{\url{https://github.com/mit-han-lab/torchsparse}.} \dotfill MIT License
\end{itemize}

\clearpage
\bibliographystyle{splncs04}
\bibliography{egbib}
\end{document}